\documentclass[10pt,a4paper,oneside]{article}

\usepackage{float}
\usepackage{amsmath,amssymb,amsfonts,amsthm}
\usepackage{mathtools}
\usepackage{bbm}
\usepackage{graphicx}
\usepackage{setspace}
\usepackage{subcaption}

\usepackage{float}
\usepackage{subfiles}
\usepackage{xr}
\usepackage{mathrsfs}
\usepackage{makeidx}

\usepackage{exscale,relsize} 
\usepackage{float}

\usepackage[toc]{appendix}

\usepackage{mleftright} 
\usepackage{theorem}
\usepackage{graphics}                 
\usepackage{graphics}                 
\usepackage{accents}
\usepackage{kbordermatrix}
\usepackage[margin=10pt,font=small,labelfont=bf]{caption}
\usepackage{tikz}
\usetikzlibrary{arrows,backgrounds,snakes}
\tikzstyle{every picture}=[scale=0.75,transform shape]
\usepackage{verbatim}
\usetikzlibrary{arrows,automata}

\usepackage{cases}
\usepackage{makeidx}
\usepackage{titlesec}
\usepackage{longtable}
\usepackage{tabu}
\usepackage{bm,xstring}

\floatstyle{ruled}
\newfloat{algorithm}{tbp}{loa}
\floatname{algorithm}{Algorithm}

\usepackage{algorithm}
\usepackage{algorithmic}
\usepackage{amsbsy}

\usepackage{adjustbox}
\usepackage{pgfplots}

\let\originalleft\left
\let\originalright\right
\renewcommand{\left}{\mathopen{}\mathclose\bgroup\originalleft}
\renewcommand{\right}{\aftergroup\egroup\originalright}

\providecommand{\U}[1]{\protect \rule{.1in}{.1in}}

\newcommand{\argmax}{\mathop{\mathrm{arg\,max}}}
\newcommand{\minimize}{\mathop{\mathrm{minimize}}}

\newcommand{\subjectto}{\mathop{\mathrm{subject\,to}}}

\newcommand{\myDelta}{{\textstyle \mathsmaller{\varDelta}}} 

\newcommand{\dsum}{\displaystyle\sum}

\mleftright 

\begin{document}

\title{A Constrained Randomized Shortest-Paths Framework for Optimal Exploration \\
 {\normalsize (draft manuscript submitted for publication and subject to changes)}
}

\author{Bertrand Lebichot$^{1}$, Guillaume Guex$^{1}$,\\ Ilkka Kivim\"aki$^{1,2}$ \& Marco Saerens$^{1,3}$
 \\
$^{1}$ICTEAM and Machine Learning Group (MLG) \\ Universit\'{e} catholique de Louvain
(UCLouvain), Belgium \\
$^{2}$Department of Computer Science \\ Aalto University, Helsinki, Finland \\
$^{3}$IRIDIA Laboratory \\ Universit\'{e} Libre de Bruxelles (ULB), Belgium
}
\maketitle

\begin{abstract}%
The present work extends the randomized shortest-paths framework (RSP), interpolating between shortest-path and random-walk routing in a network, in three directions. First, it shows how to deal with equality constraints on a subset of transition probabilities and develops a generic algorithm for solving this constrained RSP problem using Lagrangian duality. Second, it derives a surprisingly simple iterative procedure to compute the optimal, randomized, routing policy generalizing the previously developed ``soft'' Bellman-Ford algorithm. The resulting algorithm allows balancing exploitation and exploration in an optimal way by interpolating between a pure random behavior and the deterministic, optimal, policy (least-cost paths) while satisfying the constraints. Finally, the two algorithms are applied to Markov decision problems by considering the process as a constrained RSP on a bipartite state-action graph.
In this context, the derived ``soft'' value iteration algorithm appears to be closely related to dynamic policy programming \cite{Azar-2011,Azar-2012} as well as ``Kullback-Leibler" and ``path integral" control \cite{Todorov-2006,Rubin-2012,Fox-2016,Kappen-2012,Theodorou-2012,Theodorou-2013}, and similar to the reinforcement learning exploration strategy recently introduced in \cite{Asadi-2016,Asadi-2017}. This shows that this strategy is optimal in the RSP sense -- it minimizes expected path cost subject to relative entropy constraint. Simulation results on illustrative examples show that the model behaves as expected.
\end{abstract}


\section{Introduction}

\subsection{General introduction}

The present work aims to study \textbf{randomized shortest-paths} (RSP) problems with \emph{equality constraints} on the transition probabilities issued from a subset of nodes, in the context of a single source and a single destination. This extension allows to fix some transition probabilities and then finding the optimal policy which is compatible with these probabilities.
It therefore extends previous work dedicated to the RSP \cite{Saerens-2008,Yen-08K,Kivimaki-2012}, initially inspired by stochastic traffic assignment models developed in transportation science \cite{Akamatsu-1996}.

The studied problem can be described informally as follows. Our aim is to find
the optimal policy for reaching a goal node from a source node in a network by minimizing the expected cost of paths connecting these two nodes, where costs are associated to local decisions/actions. Usually, deterministic and stochastic shortest-path algorithms provide pure deterministic policies: when standing in a given state, we just choose the best path leading to minimal expected cost.
In this work, we investigate the possibility of \emph{optimally randomizing}
the policy (exploration) while fixing a subset of transition probabilities. More precisely, the agent chooses a path to the goal node within a bag of paths according to an optimal probability distribution minimizing expected cost of paths subject to a relative entropy constraint, while satisfying transition probabilities constraints on a subset of nodes. In other words, the policy is expressed in terms of paths to the goal node. Interestingly, it can be shown that this method actually defines an optimal, biased, Markov chain in which the agent is ``attracted" by the goal node (see later for details).

The degree of randomness is controlled by a \emph{temperature parameter} allowing interpolating between the least-cost solution given by the (constrained) shortest-path algorithm and a random behavior provided by a predefined, reference, random policy (a reference random walk). Randomizing the policy thus introduces a \emph{continual exploration} of the network. Standard Markov decision problems are a special case of this framework.

The originality of the work, in comparison with other models, lies in the fact that we adopt a \emph{paths-based formalism} with entropy regularization. That is, the quantities of interest are defined on the set of full paths (or trajectories) connecting the source node to the goal node in the network.
By using this paths-based formalism, as in the standard RSP \cite{Saerens-2008,Kivimaki-2012} and some models in transportation science \cite{Akamatsu-1996}, it is shown that the optimal randomized policy (both at the path level and at the edge level) can be computed by either (i) iteratively solving a system of linear equations or (ii) using a soft Bellman-Ford-like iteration algorithm.

\subsection{Why consider randomized policies?}

In practice, randomization corresponds to the
association of a probability distribution on the set of admissible
decisions in each node (\cite{Saerens-2008}, choice randomization or mixed strategy in game theory).
If no randomization is present, only the best policy is exploited.
Randomization thus appears when this probability distribution is no more
peaked on the best choice: the agent is willing to sacrifice efficiency
for exploration.
Note that randomized choices are common in a variety of fields \cite{Saerens-2008}; for instance game theory (mixed strategies; see for instance \cite{Osborne-2004}), computer sciences \cite{Motwani1995}, Markov games \cite{Littman-1994}, decision sciences \cite{Raiffa-1970}, reinforcement learning \cite{Sutton-2017}, etc.
A comprehensive related work and a detailed discussion of the reasons for randomizing the policy can be found in \cite{Saerens-2008,AchbanyNeuro-2006,Achbany-2006}, which are quickly summarized here:
\begin{itemize}
\item It is sometimes necessary to explore the environment, for instance when performing exploration in reinforcement learning \cite{Sutton-2017}.
\item If the environment is changing over time (non-stationary), the system
could benefit from randomization by performing continual exploration.
\item A deterministic policy would lead to a totally predictable behavior; on the contrary, randomness introduces unpredictability and therefore renders interception more difficult. Randomization (randomized, or mixed, strategies) has proved to be useful for this reason in game theory \cite{Osborne-2004}.
\item A randomized policy spreads the traffic over multiple paths, therefore reducing the danger of congestion. 
\item In some applications, like social networks analysis, computing a distance accounting for all paths -- and thus integrating the concept of high connectivity -- could provide better results than relying on the optimal, shortest, paths only \cite{Fouss-2016}.
\item In computer gaming, it is often desirable to be able to adapt the strength of the digital opponent \cite{Garcia-Diez-2013}. This allows modeling the behavior of incompletely rational players.
\end{itemize}

Within the context of the RSP framework, the randomness associated to paths connecting the source node and the goal node is quantified by the relative entropy, or Kullback-Leibler divergence (see, e.g., \cite{Cover-2006}), between the probability distribution defined on the paths and their likelihood according to a reference random walk on the graph -- usually following a uniform distribution on the set of available decisions. This relative entropy captures the degree of randomness of the system.
The optimal randomized policy is then obtained by minimizing the free energy -- the expected cost plus the relative entropy weighted by temperature. As already mentioned, in this work, constraints are added to the optimisation problem by considering equality constraints on some transition probabilities, which are assumed provided by the environment and which have to be verified exactly.

\subsection{Integrating constraints to the RSP framework}

Being able to deal with constraints on the transition probabilities is important in a number of applications. Indeed, we do not always have a complete control on the behavior of the system: some state transitions are intrinsically stochastic and the model has to integrate this fact. For instance, in \textbf{Markov decision processes} (MDP), part of the environment is stochastic and is modeled by a Markov chain. By the way, it will be shown that our introduced constrained randomized shortest-paths formalism subsumes simple Markov decision processes in Section \ref{Sec_BipartiteGraph01}.

Based on this constrained RSP formalism, a first, generic, algorithm for solving the constrained problem is developed by exploiting Lagrange duality.
Then, a simple, easy-to-implement, iterative algorithm, related to the ``soft" Bellman-Ford algorithm \cite{Francoisse-2013,Francoisse-2017}, is derived and its convergence to a fixed point is proved.

As an illustrative example, the framework is then used in order to solve randomized MDP problems, therefore providing a randomized policy. Markov decision processes \cite{Powell-2011,Puterman-1994,Sutton-2017,Tijms-2003}, also called stochastic shortest-path problems \cite{Bertsekas-1998,Bertsekas-2000}, are currently used in a wide range of application areas including transportation networks, medical imaging, wide-area network routing, artificial intelligence, to name a few (see, e.g., \cite{Puterman-1994,Sutton-2017,White-1985,White-1988,White-1993}).

Interestingly, when applied to MDPs, the derived Bellman-Ford-like iterative algorithm -- called here the \textbf{soft value iteration} -- is closely related to dynamic policy programming \cite{Azar-2011,Azar-2012} as well as Kullback-Leibler and path integral control \cite{Todorov-2006,Rubin-2012,Fox-2016,Kappen-2012,Theodorou-2012,Theodorou-2013}. It is also similar to the exploration strategy recently introduced in \cite{Asadi-2016,Asadi-2017}.
This shows that this proposed exploration strategy is \emph{globally optimal} in the following sense: it minimizes expected cost subject to constant relative entropy of paths probabilities when the goal state is absorbing and reachable from any other state. Interestingly, as in \cite{Francoisse-2013,Francoisse-2017} for the standard RSP without constraints, the soft value iteration algorithm extends the Bellman-Ford value iteration algorithm by simply replacing the minimum operator by a soft minimum operator.
Note that still another way of solving the problem was developed in \cite{Blaise-2013}, but this algorithm is not included here because it is less generic.

\subsection{Contributions and organization of the paper}

In brief, this work contains the following contributions:
\begin{itemize}
\item It extends randomized shortest paths to problems with constrained transition probabilities on a subset of nodes.
\item A generic algorithm solving the problem is introduced.
\item An alternative, simple and easy-to-implement, iterative algorithm for computing the optimal randomized policy is derived.
\item The constrained randomized shortest-paths framework is applied to solve standard Markov decision problems by introducing a soft value iteration algorithm.
\item Simulations on concrete problems show that the algorithms behave as expected.
\end{itemize}

As far as the organization of the paper is concerned, Section \ref{Sec_randomized_shortest_paths01}
introduces the standard randomized shortest-paths framework.  Section \ref{Sec_constrained_randomized_shortest_path01} considers randomized shortest-path problems with constraints on transition probabilities, which are then solved in Section \ref{Subsec_Lagrange_duality_edge_constraints01} by using Lagrange duality. Section \ref{Subsec_iterative_algorithm01} then develops an alternative iterative algorithm, reminiscent from the Bellman-Ford recurrence, for computing the free energy and the optimal randomized policy. In Section \ref{Sec_BipartiteGraph01}, the standard Markov decision problem is recast as a constrained randomized shortest-path problem on a bipartite graph and a soft value iteration algorithm is developed for solving it.
Section \ref{Sec_experiments01} shows some simulation examples and
Section \ref{Sec_conclusion01} is the conclusion.

\section{The standard randomized shortest-path framework}
\label{Sec_randomized_shortest_paths01}

As already stated, our formulation of the problem is based on the randomized shortest-path (RSP) framework defining, among others, a dissimilarity measure interpolating between the shortest-path distance and the commute-cost distance\footnote{On an undirected graph, the commute-cost distance appears to be proportional to the commute-time distance \cite{FoussKDE-2005,Kivimaki-2012} and to the effective resistance \cite{Chandra-1989} (also called resistance distance \cite{Klein-1993}) for a given graph -- see \cite{Fouss-2016} for a discussion.} in a graph \cite{Yen-08K,Saerens-2008,Kivimaki-2012}. The RSP framework relies on full paths instead of standard ``local" flows \cite{Ahuja-1993}.

In this section, we start by providing the necessary background and notation. Then, we proceed with a short summary of the randomized shortest-path formalism before introducing, in the next section, randomized shortest paths with constraints on the transition probabilities.

\subsection{Some background and notation}
\label{Sec_Statement}

Let us consider a weighted directed graph or network, $G$, with a set of
$n$ nodes $\mathcal{V}$ (or vertices) and a set of arcs $\mathcal{E}$ (or edges).
The graph is assumed to be \emph{strongly connected} and is represented by its $n \times n$ adjacency matrix $\mathbf{A}$, containing binary values if the graph is unweighted or non-negative, local, affinities between nodes in the case of a weighted graph.
To each edge linking node $i$ to node $j$, we also associate a
non-negative number $c_{ij}$ representing the immediate cost of following
this edge. The costs should be non-negative and are gathered in
matrix $\mathbf{C}$. Note that self-loops are forbidden; in other words, the diagonal elements of the adjacency matrix are equal to $0$. Similarly, diagonal elements of the cost are equal to $\infty$.

Moreover, a \textbf{reference random walk} (Markov chain) on $G$ is defined in the usual manner.
The choice to follow an edge from node $i$ will be made according to a
probability distribution (transition probabilities) defined
on the set $\mathcal{S}ucc(i)$ of successor nodes of $i$.
These transition probabilities, defined on each node $i$, will be denoted as
\begin{equation}
p_{ij}^{\mathrm{ref}} = \mathrm{P_{ref}} \big(s(t+1)=j|s(t)=i \big) = \dfrac{a_{ij}} {\sum_{k \in \mathcal{S}ucc(i)} a_{ik}}
\label{Eq_referenceRandomWalk01}
\end{equation}
where $a_{ij}$ is element $i,j$ of the adjacency matrix and $s(t)$ is a random variable representing the node visited by the random walker at time $t$. Furthermore, $\mathbf{P}_{\mathrm{ref}}$ will be the
matrix containing the transition probabilities $p_{ij}^{\mathrm{ref}}$
as elements. For consistency, if there is no edge between $i$ and $j$ ($a_{ij} = 0$), we
consider that $c_{ij}$ takes a large value, denoted
by $\infty$; in this case, the corresponding transition probability
must also be equal to zero, $p^{\mathrm{ref}}_{ij}=0$.

Finally, in this work, we will assume that there is a unique \emph{goal node}, which will be the last node $n$. This goal node is turned into an absorbing, killing, state in the corresponding Markov chain. Thus, any transition from this node is forbidden, that is, $p_{nj}^{\mathrm{ref}} = 0$ for all $j$ -- the random walker is killed after reaching goal state $n$.

\subsection{The standard randomized shortest-path formalism}
\label{Subsec_standard_RSP01}

The main idea behind the RSP is as follows \cite{Saerens-2008,Yen-08K,Kivimaki-2012,Francoisse-2013,Francoisse-2017}. We consider the set of all \textbf{hitting paths}, or walks, $\wp \in \mathcal{P}$ from node $1$ to the (unique) absorbing and killing node $n$ on $G$ (a bag of paths). Since the original graph is strongly connected, state $n$ can be reached from any other node of the graph. Each path $\wp$ consists in a sequence of connected nodes starting in node $1$ and ending in $n$. Then, we assign a probability distribution $\mathrm{P}(\cdot)$ (denoted as $\mathbb{P}$ for convenience) on the set of paths $\mathcal{P}$ by minimizing the relative \textbf{free energy}\footnote{Alternatively, we can adopt a maximum entropy point of view, which is equivalent when the reference probability distribution is uniform \cite{Jaakkola-2000,Jebara-2004}. Moreover, the free energy could also be defined as $\phi(\mathrm{P}) = \sum_{\wp \in \mathcal{P}} \mathrm{P}(\wp) (\tilde{c}(\wp) - c^{*}) + T \sum_{\wp \in \mathcal{P}} \mathrm{P}(\wp) \log \left( \frac{\mathrm{P}(\wp)}{\tilde{\pi}(\wp)} \right)$ where $c^{*}$ is the least cost from starting node $1$ to goal node $n$. In this case, costs are computed relatively to the shortest-path cost. This choice leads to the same probability distribution over paths (Equation (\ref{Eq_Boltzmann_probability_distribution01})).} of statistical physics \cite{Jaynes-1957,Peliti-2011,Reichl-1998},
\begin{equation}
\vline\,\begin{array}{llll}
\minimize\limits_{\{ \mathrm{P}(\wp) \}_{\wp \in \mathcal{P}}} & \phi(\mathbb{P}) = \underbracket[0.5pt][5pt]{ \dsum_{\wp \in \mathcal{P}} \mathrm{P}(\wp) \tilde{c}(\wp) }_{\text{expected cost}} + T \underbracket[0.5pt][5pt]{ \dsum_{\wp \in \mathcal{P}} \mathrm{P}(\wp) \log \left( \frac{\mathrm{P}(\wp)}{\tilde{\pi}(\wp)} \right) }_{\text{relative entropy}} \\[0.5cm]
\subjectto & \sum_{\wp\in\mathcal{P}}\textnormal{P}(\wp) = 1
\end{array}
\label{Eq_optimization_problem_BoP01}
\end{equation}
where $\tilde{c}(\wp) = \sum_{\tau = 1}^{t} c_{s(\tau-1) s(\tau)}$ is the total cumulated cost along path $\wp$ when visiting the sequence of nodes, or states, $\left( s(\tau) \right)_{\tau=0}^{t}$ and $t$ is the length of path $\wp$. Furthermore, $\tilde{\pi}(\wp) = \prod_{\tau = 1}^{t} p_{s(\tau-1) s(\tau)}^{\mathrm{ref}}$ is the product of the reference transition probabilities (see Equation (\ref{Eq_referenceRandomWalk01})) along path $\wp$ connecting node 1 to hitting node $n$ -- the likelihood of path $\wp$. It defines a \textbf{reference} probability distribution over paths as $\sum_{\wp\in\mathcal{P}} \tilde{\pi}(\wp) = 1$ \cite{Francoisse-2013,Francoisse-2017}. Note that, instead of a pure random walk, the reference probabilities $p_{ij}^{\mathrm{ref}}$ can also be set according to some prior knowledge.

The objective function in Equation (\ref{Eq_optimization_problem_BoP01}) is a mixture of two dissimilarity terms with the temperature $T$ balancing the trade-off between their relative contributions.
The first term is the expected cost for reaching goal node from source node (favoring shorter paths -- \emph{exploitation}). The second term corresponds to the relative entropy \cite{Cover-2006,Kapur-1992}, or Kullback-Leibler divergence, between the path probability distribution and the path likelihood distribution (introducing randomness -- \emph{exploration}). When the temperature $T$ is low, shorter paths are favored while when $T$ is large, paths are chosen according to their likelihood in the reference random walk on the graph $G$. Note that we should add non-negativity constraints on the path probabilities, but this is not necessary as the resulting quantities will automatically be non-negative \cite{Cover-2006,Kapur-1992}. Note that, instead of minimizing free energy, it is equivalent to minimize expected cost subject to a fixed relative entropy constraint \cite{Francoisse-2013,Fouss-2016,Francoisse-2017}.

This argument, akin to maximum entropy \cite{Jaynes-1957,Cover-2006,Kapur-1989,Kapur-1992}, leads to a \textbf{Gibbs}-\textbf{Boltzmann distribution} on the set of paths (see, e.g., \cite{Francoisse-2013,Francoisse-2017} for a detailed derivation),
\begin{equation}
\mathrm{P}^{*}(\wp) 
= \frac{\tilde{\pi}(\wp) \exp[-\theta \tilde{c}(\wp)]}{\dsum_{\wp'\in\mathcal{P}} \tilde{\pi} (\wp')\exp[-\theta \tilde{c}(\wp')]}
= \frac{\tilde{\pi}(\wp) \exp[-\theta \tilde{c}(\wp)]}{\mathcal{Z}}
\label{Eq_Boltzmann_probability_distribution01}
\end{equation}
\sloppy where $\theta = 1/T$ is the inverse temperature and the denominator $\mathcal{Z}$ $ = $ $\sum_{\wp\in\mathcal{P}} \tilde{\pi} (\wp)\exp[-\theta \tilde{c}(\wp)]$ is the \textbf{partition function} of the system.

This equation defines the \textbf{optimal randomized policy} at the \emph{paths level}, in terms of \emph{probabilities of choosing a particular path or trajectory}, $\mathrm{P}^{*}(\wp)$.
It has be shown that this set of path probabilities is exactly equivalent to the one generated by a Markov chain with biased transition probabilities $p^{*}_{ij}$ favouring shorter paths, depending on the temperature $T$ (see Equations (\ref{Eq_biased_transition_probabilities01}), (\ref{Eq_biased_transition_probabilities02}) and \cite{Saerens-2008} for details). Contrary to (\ref{Eq_Boltzmann_probability_distribution01}) defined at the path level, these transition probabilities define the optimal policy at the \emph{local, edge, level} in terms of probabilities of choosing an edge in each node. Note that a method for computing the RSP on large sparse graphs by restricting the set to paths with a finite predefined length was developed in \cite[Section 4]{Mantrach-2011}.

Several important quantities can easily be computed from this framework by, e.g., taking the partial derivative of the minimum free energy (see Equation (\ref{Eq_optimal_free_energy01}) and \cite{Yen-08K,Saerens-2008,Kivimaki-2012,Francoisse-2013,Francoisse-2017,Fouss-2016}). The quantities of interest that will be needed in this paper are introduced in the Appendix \ref{Ap_quantitiesOfInterest01}. Readers who are not familiar with the RSP framework are invited to go through this appendix before continuing the reading.

\section{Randomized shortest paths with constrained transition probabilities}
\label{Sec_constrained_randomized_shortest_path01}

Interestingly, the randomized shortest-path formulation just introduced in previous Section \ref{Subsec_standard_RSP01} can easily be extended to account for some types of constraints. The goal here will thus be to determine the best randomized policy -- the optimal transition probabilities $p^{*}_{ij}$ transporting the agent to the goal state $n$ with minimum expected cost for a given level of relative entropy, and subject to \emph{equality constraints} on some transition probabilities. We therefore have to derive the equivalent of the optimal biased transition probabilities provided by Equations (\ref{Eq_biased_transition_probabilities01}), (\ref{Eq_biased_transition_probabilities02}) in the standard RSP, but dealing now with equality constraints. This new model will be called the \textbf{constrained RSP}. As for the standard RSP, the goal node $n$ is made absorbing and killing so that all the other nodes are \emph{transient}.

As already discussed, constraints on transition probabilities are common in real-life applications where, in some (unconstrained) nodes, the agent has the control on the probability of choosing the next node while, in some other (constrained) nodes, the transition probabilities are provided by the environment and cannot be changed. An obvious example is Markov decision processes, which will be studied in the light of constrained RSP in Section \ref{Sec_BipartiteGraph01}. The constrained RSP therefore extends the range of applications of the standard RSP framework.

More concretely, we proceed as in previous section with the standard RSP, but we now constrain the transition probabilities associated to some nodes to be equal to predefined values provided by the user.
In other words, we fix the relative flow passing through the edges incident to the nodes belonging to the subset of nodes $\mathcal{C} \subset \mathcal{V} \backslash \{ n \}$ (the absorbing goal node is excluded). These nodes will be called the \textbf{constrained}, transient, nodes. The optimal transition probabilities on the remaining, unconstrained and transient, nodes (the equivalent of Equation (\ref{Eq_biased_transition_probabilities01}) to be adapted for the constrained RSP) define the optimal policy that has to be adopted by the agents at the edge level. The subset of transient, \textbf{unconstrained}, nodes will be denoted as $\mathcal{U} = \mathcal{V} \backslash ( \mathcal{C} \cup \{ n \} )$.

\subsection{The Lagrange function}

More precisely, from Equation (\ref{Eq_biased_transition_probabilities01}), the considered constraints on the nodes $i \in \mathcal{C}$ state that, on these nodes, the optimal randomized policy (transition probabilities) followed by an agent (i.e., the relative flow passing through an edge $(i,j)$) should be equal to some given values $q_{ij}$,
\begin{equation}
p^{*}_{ij}(T) = \frac{ \bar{n}_{ij}(T) }{\bar{n}_{i}(T)} = q_{ij}   \text{ for the edges starting in nodes } i \in \mathcal{C}
\label{Eq_equality_constraints_nodes01}
\end{equation}
which should be independent of the temperature $T$. Here, $\bar{n}_{i}(T)$ is the expected number of visits through node $i$ and $\bar{n}_{ij}(T)$ is the expected number of passages through edge $(i,j)$, when choosing trajectories thanks to the Gibbs-Boltzmann distribution in Equation (\ref{Eq_Boltzmann_probability_distribution01}) (see Equations (\ref{Eq_expected_visits_edges}) and (\ref{Eq_computation_node_flows01})).
The fixed transition probabilities $q_{ij}$ must be specified by the user for all the nodes in $\mathcal{C}$. Of course, we have to assume that these constraints are feasible. In particular, we must have $\sum_{j \in \mathcal{S}ucc(i)} q_{ij} = 1$ for all $i \in \mathcal{C}$ with $\mathcal{S}ucc(i)$ being the set of successor nodes of $i$.

Moreover, the RSP model (see Equation (\ref{Eq_optimization_problem_BoP01})) implies that, when $T \rightarrow \infty$, we should recover a pure random walk behavior with reference probabilities provided by Equation (\ref{Eq_referenceRandomWalk01}). Therefore, to be consistent, these reference probabilities and the $q_{ij}$ must verify $p^{\mathrm{ref}}_{ij} = p^{*}_{ij}(T = \infty) = q_{ij}$ for nodes $i \in \mathcal{C}$. Therefore the constrained transition probabilities $q_{ij}$ must be equal to the reference transition probabilities $p^{\mathrm{ref}}_{ij}$ on these constrained nodes. It will be assumed that this is the case in the sequel.

Consequently, let us consider the following Lagrange function integrating equality constraints (\ref{Eq_equality_constraints_nodes01})
\begin{align}
\mathscr{L}(\mathbb{P},\boldsymbol{\lambda})
&= \underbracket[0.5pt][5pt]{ \dsum_{\wp \in \mathcal{P}} \mathrm{P}(\wp) \tilde{c}(\wp) + T \dsum_{\wp \in \mathcal{P}} \mathrm{P}(\wp) \log \left( \frac{\mathrm{P}(\wp)}{\tilde{\pi}(\wp)} \right) }_{\text{relative free energy, }\phi(\mathbb{P})}
+ \mu \bigg( \dsum_{\wp \in \mathcal{P}} \mathrm{P}(\wp) - 1 \bigg) \nonumber \\
& + \dsum_{i \in \mathcal{C}} \dsum_{j \in \mathcal{S}ucc(i)} \lambda_{ij} \bigg[ \underbracket[0.5pt][5pt]{ \dsum_{\wp\in\mathcal{P}} \mathrm{P}(\wp) \, \eta\big((i,j) \in \wp \big) }_{\bar{n}_{ij}(T)} - q_{ij} \underbracket[0.5pt][5pt]{ \dsum_{\wp\in\mathcal{P}} \mathrm{P}(\wp) \, \eta(i \in \wp) }_{\bar{n}_{i}(T)} \bigg] 
\label{Eq_Lagrange_transition_constraints01}
\end{align}
where, as before, $\mathcal{P}$ is the set of paths connecting node $1$ to node $n$, and with $\eta\big((i,j) \in \wp\big)$ being the number of times edge $(i,j)$ appears on path $\wp$. In a similar way, $\eta\big(i \in \wp\big)$ is the number of visits to node $i$ when following path $\wp$. Therefore, the last term in the previous equation states that the constraints (\ref{Eq_equality_constraints_nodes01}) must be verified on each node $i \in \mathcal{C}$. Note that in our paths-based formalism, the expected number of visits to node $i$ is expressed by $\bar{n}_{i}(T) = \sum_{\wp\in\mathcal{P}} \mathrm{P}(\wp) \, \eta(i \in \wp)$ and the number of passages through edge $(i,j)$ by $\bar{n}_{ij}(T) = \sum_{\wp\in\mathcal{P}} \mathrm{P}(\wp) \, \eta\big((i,j) \in \wp \big)$ (see Equation (\ref{Eq_expected_visits_edges})).

Now, the Lagrange function can be rearranged as
\begin{align}
\mathscr{L}(\mathbb{P},\boldsymbol{\lambda})
&= \dsum_{\wp \in \mathcal{P}} \mathrm{P}(\wp) \bigg[ \underbracket[0.5pt][5pt]{ \tilde{c}(\wp) + \dsum_{i \in \mathcal{C}} \dsum_{j \in \mathcal{S}ucc(i)} \lambda_{ij} \, \eta\big( (i,j) \in \wp\big) - \dsum_{i \in \mathcal{C}} \eta( i \in \wp) \dsum_{j' \in \mathcal{S}ucc(i)} q_{ij'} \, \lambda_{ij'} }_{\tilde{c}'(\wp)} \bigg]
\nonumber \\
&\quad + T \dsum_{\wp \in \mathcal{P}} \mathrm{P}(\wp) \log \left( \frac{\mathrm{P}(\wp)}{\tilde{\pi}(\wp)} \right) + \mu \bigg( \dsum_{\wp \in \mathcal{P}} \mathrm{P}(\wp) - 1 \bigg) \nonumber \\
&= \dsum_{\wp \in \mathcal{P}} \mathrm{P}(\wp) \dsum_{i \in \mathcal{V}} \dsum_{j \in \mathcal{S}ucc(i)} \eta\big( (i,j) \in \wp \big) \underbracket[0.5pt][5pt]{ \bigg[ c_{ij} + \delta(i \in \mathcal{C}) \, \lambda_{ij} - \delta(i \in \mathcal{C}) \dsum_{j' \in \mathcal{S}ucc(i)} q_{ij'} \, \lambda_{ij'} \bigg] }_{\text{augmented costs } c'_{ij}}
\nonumber \\
&\quad + T \dsum_{\wp \in \mathcal{P}} \mathrm{P}(\wp) \log \left( \frac{\mathrm{P}(\wp)}{\tilde{\pi}(\wp)} \right) + \mu \bigg( \dsum_{\wp \in \mathcal{P}} \mathrm{P}(\wp) - 1 \bigg) \nonumber \\
&= \underbracket[0.5pt][5pt]{ \dsum_{\wp \in \mathcal{P}} \mathrm{P}(\wp) \tilde{c}'(\wp) + T \dsum_{\wp \in \mathcal{P}} \mathrm{P}(\wp) \log \left( \frac{\mathrm{P}(\wp)}{\tilde{\pi}(\wp)} \right) }_{\text{free energy }  \phi'(\mathbb{P}) \text{ based on augmented costs, } \tilde{c}'(\wp)}
+ \mu \bigg( \dsum_{\wp \in \mathcal{P}} \mathrm{P}(\wp) - 1 \bigg) 
\label{Eq_lagrange_function_modified_transition01}
\end{align}
where we used the Kronecker delta $\delta(i \in \mathcal{C})$ which is equal to $1$ when $i \in \mathcal{C}$ and $0$ otherwise, as well as $\eta( i \in \wp) = \sum_{j \in \mathcal{S}ucc(i)} \eta\big( (i,j) \in \wp\big)$ and $\tilde{c}(\wp) = \sum_{i \in \mathcal{V}} \sum_{j \in \mathcal{S}ucc(i)} \eta\big( (i,j) \in \wp\big) \, c_{ij}$ the total cost along path $\wp$. Thus, in (\ref{Eq_lagrange_function_modified_transition01}) the local costs $c_{ij}$ are redefined as
\begin{equation}
c'_{ij} =
\begin{cases}
c_{ij} + \underbracket[0.5pt][5pt]{ \lambda_{ij} - \dsum_{j' \in \mathcal{S}ucc(i)} q_{ij'}  \lambda_{ij'} }_{\text{extra cost $\myDelta_{ij}$}  } = c_{ij} + \myDelta_{ij} &\text{when node } i \in \mathcal{C} \\
c_{ij}  &\text{when node } i \in \mathcal{U}
\end{cases}
\label{Eq_redefined_costs_transition01}
\end{equation}
and $\mathbf{C}'$ will be the matrix containing these new costs $c'_{ij}$ where the \textbf{extra costs} are defined as $\myDelta_{ij} \triangleq \lambda_{ij} - \sum_{j' \in \mathcal{S}ucc(i)} q_{ij'}  \lambda_{ij'}$. 

These new costs $c'_{ij}$, augmented by the extra costs coming from the Lagrange mutipliers, will be called the \textbf{augmented costs}. We observe that Equation (\ref{Eq_lagrange_function_modified_transition01}) is exactly a randomized shortest-paths problem (see Equation (\ref{Eq_optimization_problem_BoP01})) containing augmented costs instead of the initial costs, which can be solved by a standard RSP algorithm.

We further observe that the weighted (by transition probabilities) means of the extra costs must be equal to zero on each node $i \in \mathcal{C}$:
\begin{equation}
\sum_{j \in \mathcal{S}ucc(i)} q_{ij} \myDelta_{ij} = 0 \quad \text{for each } i \in \mathcal{C}
\label{Eq_cost_updates_sum_to_one01}
\end{equation}
In other words, the extra costs are centered with respect to the weights $q_{ij}$ on each constrained node. Interestingly, this implies that the weighted average of the augmented costs is equal to the weighted average of the original costs on each constrained node $i$, $\sum_{j \in \mathcal{S}ucc(i)} q_{ij} c'_{ij} = \sum_{j \in \mathcal{S}ucc(i)} q_{ij} c_{ij}$.
In this case, the perceived cost (cost plus extra cost) when visiting any node using the augmented costs is exactly the same in average as the perceived real cost (cost only) as in the case where no constraint is introduced.

Thus, in Equation (\ref{Eq_lagrange_function_modified_transition01}), everything happens as if the costs have been redefined by taking into account the Lagrange parameters. The extra costs, depending on these Lagrange parameters, can therefore be interpreted as extra virtual costs necessary to exactly satisfy the equality constraints, in the same way as when considering the dual problem in linear programming \cite{Griva-2008}.

Let $\phi'(\mathbb{P}) = \sum_{\wp \in \mathcal{P}} \mathrm{P}(\wp) \tilde{c}'(\wp) + T \sum_{\wp \in \mathcal{P}} \mathrm{P}(\wp) \log \left( \frac{\mathrm{P}(\wp)}{\tilde{\pi}(\wp)} \right)$ be the relative free energy obtained from these augmented costs (see Equation (\ref{Eq_lagrange_function_modified_transition01})).
We now address the problem of computing the Lagrange parameters $\lambda_{ij}$ and the extra costs $\myDelta_{ij}$ by Lagrange duality.


\section{Solving constrained RSP problems by Lagrange duality}
\label{Subsec_Lagrange_duality_edge_constraints01}

In this section, we will take advantage of the fact that, in our formulation of the problem, the Lagrange dual function and its gradient with respect to a set of Lagrange parameters associated to a node are easy to compute. Indeed, the situation is equivalent to maximum entropy problems under constraints (see, e.g., \cite{Jaakkola-2000,Jebara-2004}), so that the same methodology can be used for optimising the objective function. This will provide a \emph{generic algorithm} for solving constrained RSP problems based on Lagrange duality.

As the objective function is convex and all the equality constraints are linear, there is only one global minimum and the duality gap is zero \cite{Bertsekas-1999,Culioli-2012,Griva-2008}. The optimum is a saddle point of the Lagrange function and a common optimization procedure (\cite{Bertsekas-1999,Culioli-2012,Griva-2008}, related to the Arrow-Hurwicz-Uzawa method \cite{Arrow-1958}) consists in sequentially (i) solving the primal while considering the Lagrange parameters as fixed, which provides the dual Lagrange function $\mathscr{L}^{*}(\boldsymbol{\lambda})$, and then (ii) optimizing the obtained dual Lagrange function (which is concave) with respect to a subset of Lagrange parameters (a block $\mathcal{B}$) until convergence. 

In our context, this provides the two following steps \cite{Griva-2008}, which are computed iteratively on blocks of Lagrange parameters $\mathcal{B}$,
\begin{equation}
\begin{cases}
\mathscr{L}^{*}(\boldsymbol{\lambda}^{(t)}) = \min \limits_{\mathbb{P} \equiv \{ \mathrm{P}(\wp) \}_{\wp \in \mathcal{P}}} \mathscr{L}(\mathbb{P},\boldsymbol{\lambda}^{(t)})  &\text{\footnotesize (compute the dual Lagrange function)} \\
\lambda_{ij}^{(t+1)} = \argmax\limits_{\lambda_{ij}^{(t)} \in \mathcal{B}^{(t)}} \mathscr{L}^{*}(\boldsymbol{\lambda}^{(t)}) \text{ for } \lambda_{ij}^{(t)} \in \mathcal{B}^{(t)} &\text{\footnotesize (maximize the dual Lagrange function)} \\
\lambda_{ij}^{(t+1)} = \lambda_{ij}^{(t)} \text{ for } \lambda_{ij}^{(t)} \notin \mathcal{B}^{(t)} &\text{\footnotesize (keep the other Lagrange parameters)}
\end{cases}
\label{Eq_primal_dual_lagrangian01}
\end{equation}
and the first maximization is performed subject to non-negativity and sum-to-one constraints. This is the procedure that will be followed, where the dual function will be maximized through a simple block coordinate ascend on Lagrange parameters. Each block at a given step $t$ of the iteration will contain the Lagrange parameters associated to the node $i$ processed at that time step (the edges incident to node $i$, $\mathcal{B}^{(t)} = \mathcal{S}ucc(i)$) and the procedure is iterated on the set of constrained nodes ($i \in \mathcal{C}$).

\subsection{Computing the dual Lagrange function}

We already know from (\ref{Eq_Boltzmann_probability_distribution01}) that in the first step in Equation (\ref{Eq_primal_dual_lagrangian01}) the optimal probability distribution is obtained with
\begin{equation}
\mathrm{P}^{*}(\wp) 
= \frac{\tilde{\pi}(\wp) \exp[-\theta \tilde{c}'(\wp)]}{\dsum_{\wp'\in\mathcal{P}} \tilde{\pi} (\wp')\exp[-\theta \tilde{c}'(\wp')]}
= \frac{\tilde{\pi}(\wp) \exp[-\theta \tilde{c}'(\wp)]}{\mathcal{Z}'}
\label{Eq_Boltzmann_probability_distribution_extended01}
\end{equation}
where $\tilde{c}'(\wp)$ is the augmented cost of path $\wp$.


Then, from Equations (\ref{Eq_optimal_free_energy01}) and (\ref{Eq_lagrange_function_modified_transition01}), the dual Lagrange function can easily be computed in function of the partition function defined from the augmented costs \cite{Jebara-2004},
\begin{equation}
\mathscr{L}^{*}(\boldsymbol{\lambda}) = -T \log \mathcal{Z}'
\label{Eq_dual_lagrangian_transition_constraints01}
\end{equation}
and will be maximized at each time step with respect to the $\{ \lambda_{ij} \}$ with $i \in \mathcal{C}$ and $j \in \mathcal{S}ucc(i)$. In addition, by extension of Equation (\ref{Eq_optimal_free_energy01}) to any transient nodes (see Equation (\ref{Eq_free_energy_definition01})), the minimum free energy from any node $i$ (see \cite{Kivimaki-2012,Francoisse-2013,Francoisse-2017} for details) is given by
\begin{equation}
\phi^{*}_{i} = -T \log z_{in} = -\tfrac{1}{\theta} \log z_{in}
\label{Eq_minimum_free_energy01}
\end{equation}
where the backward variable $z_{in}$ (element $i$, $n$, of the fundamental matrix $\mathbf{Z}$, see Equation (\ref{Eq_forward_backward_variables01})) is now computed from the \emph{augmented costs}.

\subsection{Maximizing the dual Lagrange function}

Let us now maximize the dual function by using a simple block coordinate ascend.
Because $\bar{n}_{i}(T) = \sum_{j \in \mathcal{S}ucc(i)} \bar{n}_{ij}(T)$, by following the reasoning of previous subsection (see Equation (\ref{Eq_expected_visits_edges})), we obtain for constrained nodes $i \in \mathcal{C}$
\begin{align}
\frac{\partial \mathscr{L}^{*}(\boldsymbol{\lambda})}{\partial \lambda_{ij}} 
&= \frac{\partial (-T \log \mathcal{Z}')} {\partial \lambda_{ij}}
= \sum_{j' \in \mathcal{S}ucc(i)} \frac{\partial (-T \log \mathcal{Z}')} {\partial c'_{ij'}} \frac{\partial c'_{ij'}} {\partial \lambda_{ij}} \nonumber \\
&= \sum_{j' \in \mathcal{S}ucc(i)} \bar{n}_{ij'}(T) (\delta_{jj'} - q_{ij})
= \bar{n}_{ij}(T) - q_{ij} \bar{n}_{i}(T)
\label{Eq_dual_function_derivative_node_constraints02}
\end{align}

Quite naturally, and similarly to maximum entropy problems \cite{Kapur-1992}, setting the result to zero provides the constraints on nodes $i \in \mathcal{C}$,
\begin{equation}
\frac{\bar{n}_{ij}(T)}{\bar{n}_{i}(T)} = q_{ij}
\label{Eq_constraint_imposed01}
\end{equation}
and we now have to solve these equations in terms of the Lagrange parameter $\lambda_{ij}$.

\begin{algorithm}[t!]
\caption[Algorithm01]
{Computing the optimal randomized policy of a constrained RSP problem.}
\algsetup{indent=2em, linenodelimiter=.}
\begin{algorithmic}[1]
\small
\REQUIRE $\,$ \\
 -- The $n\times n$ adjacency matrix $\mathbf{A}$ of a strongly connected directed graph, containing non-negative edge affinities. Node 1 is the starting node and node $n$ the goal node.\\
 -- The $n\times n$ cost matrix $\mathbf{C}$ of the graph, containing non-negative edge costs.\\
 -- The set of unconstrained nodes $\mathcal{U}$ and constrained nodes $\mathcal{C}$.\\
 -- The positive inverse temperature parameter $\theta$. \\
 
\ENSURE $\,$ \\
 -- The $(n-1) \times n$ matrix $\mathbf{P}^{*}$ containing optimal transition probabilities (the policy). \\

~\\

\STATE $\mathbf{D} \leftarrow \mathbf{Diag}(\mathbf{A}\mathbf{e})$ \COMMENT{the diagonal out-degree matrix; $\mathbf{e}$ is a vector full of $1$'s} \\
\STATE $\mathbf{P}_\mathrm{ref} \leftarrow \mathbf{D}^{-1} \mathbf{A}$ \COMMENT{the $n \times n$ reference transition probabilities matrix} \\
\STATE $\mathbf{C}' \leftarrow \mathbf{C}$ \COMMENT{initialise the augmented costs matrix} \\
\STATE Set row $n$ of $\mathbf{P}_\mathrm{ref}$ to $\boldsymbol{0}^{\mathrm{T}}$  \COMMENT{row $n$ set to zero: node $n$ is made absorbing and killing} \\
\REPEAT[main iteration loop]
\FOR[loop on constrained nodes in $\mathcal{C}$]{\textbf{each} $i \in \mathcal{C}$}
\STATE $\mathbf{W} \leftarrow \mathbf{P}_{\mathrm{ref}} \circ \exp[-\theta \mathbf{C}']$ \COMMENT{compute the auxiliary matrix $\mathbf{W}$ in terms of current augmented costs; $\circ$ is the elementwise matrix product} \\
\STATE Solve $(\mathbf{I} - \mathbf{W}) \mathbf{z}_{\mathrm{b}} = \mathbf{e}_{n}$ with respect to $\mathbf{z}_{\mathrm{b}}$ \COMMENT{compute the backward variable $\mathbf{z}_{\mathrm{b}} = \mathbf{Z} \mathbf{e}_{n}$ (column $n$ of the fundamental matrix $\mathbf{Z} = (\mathbf{I} - \mathbf{W})^{-1}$) where $\mathbf{e}_{n}$ is a vector full of $0$'s except element $n$ which is equal to $1$}
\STATE $\bm{\phi}^{*} \leftarrow - \tfrac{1}{\theta} \log \mathbf{z}_{\mathrm{b}}$ and then $ \phi^{*}_{n} \leftarrow 0$ \COMMENT{elementwise natural logarithm: compute the vector of free energies, and force $0$ on the goal node $n$} \\
\FOR[update the augmented costs on edges incident to constrained node $i$]{\textbf{each} $j \in \mathcal{S}ucc(i)$}
\STATE $ c'_{ij} \leftarrow - \phi^{*}_{j} + \sum_{k \in \mathcal{S}ucc(i))} p^{\mathrm{ref}}_{ik} (c_{ik} + \phi^{*}_{k})$ \COMMENT{augmented cost update for edge $(i,j)$} \\
\ENDFOR
\ENDFOR
\UNTIL{convergence of the free energy vector}
\STATE $\mathbf{Q} \leftarrow \mathbf{P}_{\mathrm{ref}} \circ \exp[-\theta ( \mathbf{C}' + \mathbf{e} (\bm{\phi}^{*})^{\mathrm{T}} )]$ \COMMENT{compute the numerator of the optimal transition probabilities matrix} \\
\STATE Remove row $n$ of matrix $\mathbf{Q}$ \COMMENT{delete the zero row corresponding to the absorbing, goal, node $n$} \\
\STATE $\mathbf{s} \leftarrow \mathbf{Q} \mathbf{e}$ \COMMENT{the row sums vector for normalization} \\
\STATE $\mathbf{P}^{*} \leftarrow \mathbf{Q} \div \big( \mathbf{s} \mathbf{e}^{\mathrm{T}} \big)$ \COMMENT{the $(n-1) \times n$ optimal transition probabilities matrix (the policy); $\div$ is the elementwise division. We divide each row of $\mathbf{Q}$ by its sum.} \\
\RETURN $\mathbf{P}^{*}$
\end{algorithmic}
\label{Algorithm01} 
\end{algorithm}


\subsection{Computing the Lagrange parameters and the augmented costs}
\label{Subsec_augmented_costs_computation_free_energy01}

Recalling that $\bar{n}_{i}(T) = \sum_{j \in \mathcal{S}ucc(i)} \bar{n}_{ij}(T)$ and Equations (\ref{Eq_computation_edge_flows01})-(\ref{Eq_computation_node_flows01}), we obtain by imposing the constraint (\ref{Eq_constraint_imposed01}) for a node $i \in \mathcal{C}$ and $j \in \mathcal{S}ucc(i)$,
\begin{align}
&\frac{ p^{\mathrm{ref}}_{ij} \exp[-\theta c'_{ij}]  z_{jn} } { \sum_{j' \in \mathcal{S}ucc(i)} p^{\mathrm{ref}}_{ij'} \exp[-\theta c'_{ij'}]  z_{j'n} } \nonumber \\
&\qquad =\frac{ p^{\mathrm{ref}}_{ij} z_{jn} \exp[-\theta c_{ij}] \exp[-\theta \myDelta_{ij}] } { \sum_{j' \in \mathcal{S}ucc(i)} p^{\mathrm{ref}}_{ij'} z_{j'n} \exp[-\theta c_{ij'}] \exp[-\theta \myDelta_{ij'}] }
= q_{ij}
\label{Eq_to_solve_augmented_costs01}
\end{align}
The goal now is to compute the new augmented cost (and thus the new extra costs $\myDelta_{ij}$ and the Lagrange parameters $\lambda_{ij}$, see Equation (\ref{Eq_redefined_costs_transition01})) corresponding to node $i \in \mathcal{C}$ by isolating the $\myDelta_{ij}$ with $j \in \mathcal{S}ucc(i)$ in the previous Equation (\ref{Eq_to_solve_augmented_costs01}). In Appendix \ref{Appendix0}, it is shown that we obtain (see Equation (\ref{Eq_my_delta_expression_appendix01}))
\begin{equation}
\myDelta_{ij} = -(c_{ij} + \phi^{*}_{j}) + \sum_{k \in \mathcal{S}ucc(i)} p^{\mathrm{ref}}_{ik} (c_{ik} + \phi^{*}_{k}), \text{ for each } j \in \mathcal{S}ucc(i)
\label{Eq_my_delta_expression02}
\end{equation}
which allows to directly compute the new augmented costs
\begin{equation}
c_{ij}' = c_{ij} + \myDelta_{ij} = \sum_{k \in \mathcal{S}ucc(i)} p^{\mathrm{ref}}_{ik} (c_{ik} + \phi^{*}_{k}) - \phi^{*}_{j}, \text{ for each } j \in \mathcal{S}ucc(i)
\label{Eq_formula_augmented_costs_free_energy01}
\end{equation}
and, after convergence, this expression must be exactly verified by the augmented costs on all the constrained nodes.

Equation (\ref{Eq_formula_augmented_costs_free_energy01}) suggests the following updating rule (bloc coordinate ascend) to be applied on all the edges incident to $i$ at each iteration
\begin{equation}
c_{ij}' \leftarrow \sum_{k \in \mathcal{S}ucc(i)} p^{\mathrm{ref}}_{ik} (c_{ik} + \phi^{*}_{k}) - \phi^{*}_{j}, \text{ for each } j \in \mathcal{S}ucc(i)
\label{Eq_redefined_costs_update_transition01}
\end{equation}
to be repeated on all constrained nodes (one constrained node $i$ processed at each iteration step) until convergence.

Moreover, it can easily be shown from the previous results that the Lagrange multipliers are given\footnote{Up to the addition of a constant, as they must be centered.} by
\begin{equation}
\lambda_{ij} = -(c_{ij} + \phi^{*}_{j})
\end{equation}

Let us now summarize the whole procedure.

\subsection{The complete procedure}
\label{Subsec_constrained_RSP_procedure01}

Therefore, after specifying a parameter $\theta$ and initializing the augmented costs $c_{ij}'$ to the real costs $c_{ij}$, the final block coordinate ascend procedure iterates the following steps for updating the augmented costs associated to a constrained node $i$:
\begin{enumerate}
  \item The elements of the fundamental matrix are computed thanks to Equation (\ref{Eq_fundamentalMatrix01}) from the \emph{current augmented costs} $c_{ij}'$ (defined in Equation (\ref{Eq_redefined_costs_transition01})) and from the transition matrix of the natural random walk on $G$ (Equation (\ref{Eq_referenceRandomWalk01})), where goal node $n$ is made absorbing and killing, $\mathbf{Z} = (\mathbf{I} - \mathbf{W})^{-1}$ with $\mathbf{W} = \mathbf{P}_{\mathrm{ref}} \circ \exp[-\theta \mathbf{C}']$.
  \item Compute the minimum free energies on node $i$ and its adjacent nodes ($j \in \mathcal{S}ucc(i)$) thanks to Equation (\ref{Eq_minimum_free_energy01}), $\phi^{*}_{i} = -\tfrac{1}{\theta} \log z_{in}$.
  \item The augmented costs are updated on all edges incident to node $i$ ($j \in \mathcal{S}ucc(i)$) thanks to Equation (\ref{Eq_redefined_costs_update_transition01}), $c_{ij}' \leftarrow \sum_{k \in \mathcal{S}ucc(i))} p^{\mathrm{ref}}_{ik} (c_{ik} + \phi^{*}_{k}) - \phi^{*}_{j}$. Then, go back to step 1 and proceed with another constrained node $i$.
\end{enumerate}

The previous steps are thus performed repeatedly on the constrained nodes $i \in \mathcal{C}$ and the whole procedure is iterated until convergence. Then, the optimal policy is obtained from Equation (\ref{Eq_biased_transition_probabilities02}) by using the augmented costs $c'_{ij}$ instead of $c_{ij}$ (also for computing the backward variables $z_{in}$). This provides the optimal transition probabilities $p^{*}_{ij}(T)$ on the unconstrained nodes -- for the constrained nodes, the optimal transition probabilities are of course equal to the reference transition probabilities.
The resulting algorithm is presented in Algorithm \ref{Algorithm01}. Note that in this algorithm (line 8), instead of computing the fundamental matrix $\mathbf{Z}$, we prefer to simply calculate the backward variables vector $\mathbf{z}_{\mathrm{b}} = \mathbf{Z} \mathbf{e}_{n}$ containing the elements $z_{in}$.

Let us now present an alternative, iterative, procedure, reminiscent of the Bellman-Ford formula for finding the shortest-path distance in a graph and the value iteration in Markov decision problems, solving the constrained RSP problem.


\section{Solving constrained RSP problems by a simple iterative algorithm}
\label{Subsec_iterative_algorithm01}

This section introduces an alternative way of solving constrained randomized shortest-paths problems, based on an extension of Equation (\ref{Eq_potential_recurrence_formula01}) computing the free energy from each transient node to the goal node \cite{Francoisse-2013,Francoisse-2017}. Once the free energy has been computed for all nodes, the optimal policy is easily obtained by the closed-form expression (\ref{Eq_biased_transition_probabilities02}).

\subsection{An optimality condition in terms of free energy}

Recall that the quantity $\phi^{*}_{i}(T) = -\frac{1}{\theta} \log z_{in}$ with $\theta = 1/T$ (see Equation (\ref{Eq_minimum_free_energy01})), where $z_{in}$ is the backward variable introduced in Equation (\ref{Eq_forward_backward_variables01}), is called the (minimum) relative, directed, free energy potential\footnote{Often simply called the free energy.} of the constrained RSP system associated to the different nodes $i \in \mathcal{V}$. As before, the dependence of the free energy on $T$ will be omitted.

Inspired by the standard bag-of-paths framework \cite{Francoisse-2013,Francoisse-2017}, it is shown in Appendix \ref{Appendix1} that, at optimality, the recurrence relations computing the minimal free energy of the constrained RSP system are of the following form
\begin{equation}
\phi^{*}_{i} =
  \begin{cases}
   -\frac{1}{\theta} \log \Bigg[ {\displaystyle \sum_{j \in \mathcal{S}ucc(i)}} p_{ij}^{\mathrm{ref}} \exp \Big[-\theta \big( c_{ij} + \phi^{*}_{j} \big) \Big] \Bigg] & \text{if } i \in \mathcal{U} \\
   {\displaystyle \sum_{j \in \mathcal{S}ucc(i)}} p_{ij}^{\mathrm{ref}} \big( c_{ij} + \phi^{*}_{j} \big) & \text{if } i \in \mathcal{C} \\
   \phantom{-} 0       & \text{if } i=n
  \end{cases}
\label{Eq_value_iteration_constrained_RSP01}
\end{equation}
where, as usual, $\mathcal{S}ucc(i)$ is the set of successor nodes of node $i$ in the network and $\mathcal{U}$, $\mathcal{C}$ are resectively the sets of unconstrained and constrained nodes. This equation states the necessary optimality conditions for the constrained RSP in terms of the free energy. The first line of this equation is simply the optimality condition previously obtained for the standard RSP (see Equation (\ref{Eq_potential_recurrence_formula01}) or \cite{Francoisse-2013,Francoisse-2017}), which should apply on unconstrained nodes. The second line also makes sense as it corresponds to the recurrence expression for computing expected cost for reaching goal node $n$ from constrained node $i$ (transition probabilities are fixed on these nodes) \cite{Kemeny-1960,Norris-1997,Taylor-1998}.

\subsection{Computing the randomized policy}

The previous Equation (\ref{Eq_value_iteration_constrained_RSP01}) suggests a simple fixed-point iteration algorithm for computing the solution of the constrained RSP by replacing the equality ``$=$" by an update ``$\leftarrow$". The update is iterated until convergence to a fixed point, in the same way as the value iteration algorithm in Markov decision processes, eventually providing the values of the free energy on each node.
Then, the optimal, local, randomized policy can be obtained by Equation (\ref{Eq_biased_transition_probabilities02}) for unconstrained nodes $i \in \mathcal{U}$.
For constrained nodes, the transition probabilities are of course fixed to $p_{ij}^{\mathrm{ref}} = q_{ij}$.

In \cite{Tahbaz-2006}, it was shown that the iterative update of an expression similar (but somewhat simpler) to Equation (\ref{Eq_potential_recurrence_formula01}) converges and its limit is independent of the initial values. We prove the same property for the iteration of Equation (\ref{Eq_value_iteration_constrained_RSP01}) in Appendix \ref{Appendix2} by using a fixed-point theorem point of view, showing that the update of (\ref{Eq_value_iteration_constrained_RSP01}) is a contraction mapping.
Besides theoretical convergence, we observed empirically in all our experiments that both techniques (the iterative and the generic constrained RSP procedures) converge and provide exactly the same policies.

\section{Markov decision processes as a constrained RSP on a bipartite graph}
\label{Sec_BipartiteGraph01}

The previous sections developed all the needed tools for computing
an optimal randomized policy on a Markov decision process (MDP), which is done in this section.

Recall that, as in \cite{Bertsekas-2000}, we assume that there
is a special cost-free goal state
$n_{\mathcal{S}}$; once the system has reached that state, it simply disappears (killing state --
state $n_{\mathcal{S}}$ has no outgoing link).
As in \cite{Saerens-2008}, we will also consider a problem structure such that termination
is inevitable. Thus, the horizon is in effect finite, but its length
is random and it depends on the policy being used. The conditions
for which this is true are, basically, related to the fact that the
goal state can be reached in a finite number of steps from any
potential initial state; for a rigorous treatment, see e.g. \cite{Bertsekas-2000,
Bertsekas-1996}.

The main objective is thus, as before, to design a randomized policy
minimizing the expected cost-to-go subject
to an (relative) entropy constraint controlling the total randomness spread in
the Markov process, and therefore the exploration effort. In other words,
we are looking for an optimal policy or, in our case, the optimal
transition probabilities matrix of a finite,
first-order, Markov chain minimizing the expected cost needed
to reach the goal state from the initial state, while fixing the
entropy spread in the chain as well as the transition probabilities
provided by the environment.

Therefore, the solution is obtained by the algorithms described in Sections
\ref{Subsec_Lagrange_duality_edge_constraints01} and \ref{Subsec_iterative_algorithm01}, solving the constrained RSP, applied to a bipartite graph, as described now.

\begin{figure}[t]
\begin{center}
      \includegraphics[width = 0.50\textwidth, trim=0cm 0cm 0cm 0cm, clip=true]{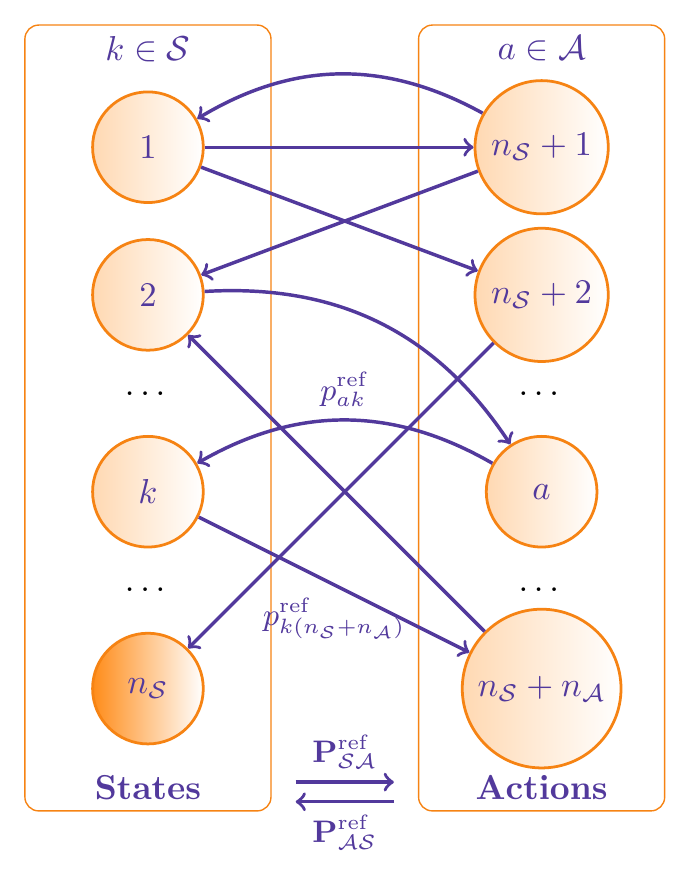}
\end{center}
\caption{A simple Markov decision process modeled as a bipartite graph $G_{\mathrm{b}}$ with states on the left side ($\mathcal{S}$) and control actions on the right ($\mathcal{A}$). Node 1 is the initial state while node $n_{\mathcal{S}}$ is the absorbing, goal, state of the process. The reference transition probabilities from states to actions $p^{\mathrm{ref}}_{ka}$ (the reference policy) are gathered in matrix $\mathbf{P}_{\mathcal{S} \mathcal{A}}^{\mathrm{ref}}$ while the transition probabilities from actions to states $p^{\mathrm{ref}}_{ak}$, provided by the environment, are gathered in matrix $\mathbf{P}_{\mathcal{A} \mathcal{S}}^{\mathrm{ref}}$.}
\label{Fig_bipartiteGraph01}
\end{figure}

\subsection{The basic model}
\label{Subsec_basic_MDP_model01}

The Markov decision process is now viewed as a constrained randomized shortest paths problem on a bipartite graph (see Figure \ref{Fig_bipartiteGraph01}). Let us first describe the structure of this bipartite graph. Then, we examine how the reference transition probabilities, corresponding to the natural random walk on this graph, are defined. Finally, the way to compute the optimal randomized policy is detailed.

\subsubsection{Definition of the bipartite graph}

The process can be modeled as a directed \textbf{bipartite graph} $G_{\mathrm{b}}$ (Figure \ref{Fig_bipartiteGraph01}) in which the \emph{left nodes} are the original states $\mathcal{S}$ and the \emph{right nodes} correspond to the possible actions associated to the states, $\mathcal{A} = \mathcal{A}(1) \cup \mathcal{A}(2) \cup \ldots \cup \mathcal{A}(n_{\mathcal{S}}-1)$ where $\mathcal{A}(k)$ is the set of actions available in state $k$. Note that the last, goal, state $n_{\mathcal{S}}$ is absorbing and has no associated action. We thus have $n_{\mathcal{S}} = |\mathcal{S}|$ left nodes (called \emph{states} or \emph{state nodes}) and $n_{\mathcal{A}} = |\mathcal{A}|$ right nodes (called \emph{actions} or \emph{action nodes}). 

Note that each action associated to a state is a node of $G_{\mathrm{b}}$, even if the same action is also available in some other states. In other words, action nodes are duplicated for each state in which they appear. Therefore, the number of such right states is $|\mathcal{A}| = |\mathcal{A}(1)| + |\mathcal{A}(2)| + \cdots + |\mathcal{A}(n_{\mathcal{S}}-1)| = n_{\mathcal{A}}$.

Moreover, it is assumed that, in this bipartite graph $G_{\mathrm{b}}$, the nodes corresponding to states $\mathcal{S}$ are numbered first (from $1$ to $n_{\mathcal{S}}$) and actions $\mathcal{A}$ are following (from $n_{\mathcal{S}}+1$ to $n_{\mathcal{S}}+n_{\mathcal{A}}$). Moreover, the set of available actions in any state $k$ is nothing else that the successor nodes of $k$ in $G_{\mathrm{b}}$, $\mathcal{A}(k) = \mathcal{S}ucc(k)$.

\subsubsection{Defining reference probabilities on the bipartite graph}

We will now describe how the reference transition probabilities (see Equation (\ref{Eq_referenceRandomWalk01}))) as well as the constrained nodes are assigned on our graph $G_{\mathrm{b}}$. In the case of a \emph{pure, natural, random walk} on $G_{\mathrm{b}}$, corresponding to $T \rightarrow \infty$ in Equation (\ref{Eq_optimization_problem_BoP01})), we consider that agents are sent from the initial state $1$ and that, at each state $s=k$ ($n_{\mathcal{S}}$ states in total), they choose an action $a$ with probability mass $p^{\mathrm{ref}}_{ka}$, $k \in \mathcal{S}$ and $a \in \mathcal{A}(k)$. When no prior information on the system is available, these are usually set to $p^{\mathrm{ref}}_{ka} = 1/|\mathcal{A}(k)|$, a uniform distribution.
Agents in state $k$ then jump to some action node $a$ with probability $p^{\mathrm{ref}}_{ka}$, meaning that they perform the action $a$ and incur a finite cost $c_{ka}$ associated to the execution of action $a$ in state $k$.

Furthermore, the agent then moves from action node $a$ to the next state $s=l$ with a reference transition probability $p^{\mathrm{ref}}_{al}$ provided by the environment as in standard Markov decision processes, where $l \in \mathcal{S}$, depending on the chosen action. These transition probabilities from action nodes to state nodes cannot be controlled or changed, and correspond therefore to the \emph{constrained} transition probabilities, $q_{al}$, as discussed in the previous section describing the constrained RSP.

Thus, in our bipartite graph $G_{\mathrm{b}}$, \emph{the set of state nodes $\mathcal{S}$ is nothing else than the set of unconstrained nodes} $\mathcal{U}$, together with the absorbing, goal, node $n_{\mathcal{S}}$, in the constrained RSP framework. Conversely, \emph{the set of action nodes $\mathcal{A}$ corresponds exactly to the constrained nodes} $\mathcal{C}$ because the transition probabilities are fixed by the environment.
Consequently, the transition and the cost matrices defined on the bipartite graph $G_{\mathrm{b}}$ are
\begin{equation}
\mathbf{P}_{\mathrm{ref}} = \kbordermatrix{
      &  \mathcal{S}  &  \mathcal{A}  \cr
\mathcal{S} &  \mathbf{O}  &  \mathbf{P}^{\mathrm{ref}}_{\mathcal{S}\mathcal{A}}  \cr
\mathcal{A} &  \mathbf{P}^{\mathrm{ref}}_{\mathcal{A}\mathcal{S}}  &  \mathbf{O}  \cr
},
\quad
\mathbf{C}_{\mathrm{b}} = \kbordermatrix{
      &  \mathcal{S}  &  \mathcal{A}  \cr
\mathcal{S} &  \mathbf{O}  &  \mathbf{C}  \cr
\mathcal{A} &  \mathbf{O}  &  \mathbf{O}  \cr
},
\label{Eq_block_transition_cost_matrix01}
\end{equation}
where $\mathbf{O}$ is a $0$ matrix of the appropriate size.

Note that, as for standard Markov decision processes, it is assumed that there is a non-negative cost associated to the transitions between state nodes and action nodes (the cost of choosing the action in the state), while no cost is associated to the transitions between action nodes and state nodes\footnote{Note that an additional cost could also be assigned to the transition to state node, after action $a$ is performed, as, e.g., in \cite{Sutton-2017}, but in this work we adopt the simpler setting where the cost is a function of the action $a$ only. However, our algorithm can straightforwardly be adapted to costs on actions-to-states \cite{Powell-2011,Sutton-2017,Tijms-2003}.}.

\subsubsection{Computing the optimal randomized policy}

Now that the bipartite graph $G_{\mathrm{b}}$ is defined, solving the MDP problem simply aims at applying the constrained RSP procedure defined in the Section \ref{Subsec_Lagrange_duality_edge_constraints01} on $G_{\mathrm{b}}$ (see Algorithm \ref{Algorithm01}).
This procedure returns matrix $\mathbf{P}^{*}$, containing the optimal randomized policy $p^{*}_{ka}(T)$ for each state node $k$. More precisely, the elements $\{p^{*}_{ka}(T): (k \in \mathcal{S}) \land (a \in \mathcal{A}(k)) \}$ contain, for each state $k$, an optimal probability distribution on the set $\mathcal{A}(k)$ of actions available in this state, provided by Equation (\ref{Eq_biased_transition_probabilities02}), and gradually biasing the walk towards the optimal, deterministic, policy when temperature is low. Indeed, when the temperature $T$ decreases, the agents are more and more exploiting good policies while still exploring the environment -- they interpolate between a purely random behavior (guided by the reference probabilities) and the best, deterministic, policy solving the Markov decision process, provided, e.g., by the well-known value iteration algorithm \cite{Puterman-1994,Sutton-2017,Bertsekas-1998,Bertsekas-2000}.
This policy is optimal in the sense that it minimizes expected cost for a given degree of relative entropy (see Equation (\ref{Eq_optimization_problem_BoP01})).
%

In summary, the MDP problem tackled in this section simply corresponds to a constrained randomized shortest-path problem (RSP) on $G_{\mathrm{b}}$.
We now describe a more direct way for obtaining the optimal randomized policy avoiding the construction of $G_{\mathrm{b}}$, and inspired by the value iteration algorithm. It is derived as a special case of the iterative procedure for solving constrained RSP problems developed in Section \ref{Subsec_iterative_algorithm01}.

\subsection{A soft value iteration algorithm}
\label{Subsec_value_iteration01}

Interestingly and surprisingly, we will show in this section that, as for the standard RSP (see Equation (\ref{Eq_potential_recurrence_formula01}) and its discussion below), replacing the minimum operator by a softmin operator (\ref{Eq_softmin01}) in the standard value iteration algorithm recovers exactly the iterative procedure solving the constrained RSP of Section \ref{Subsec_iterative_algorithm01} -- and providing an optimal randomized policy in the RSP sense to our Markov decision problem. This was already observed in the context of the standard RSP where we obtained a randomized Bellman-Ford recurrence expression where the min operator is replaced by a softmin operator \cite{Francoisse-2013,Francoisse-2017}.

This implies that the recent propositions of using the softmin function for exploration in reinforcement learning \cite{Azar-2011,Azar-2012,Asadi-2016,Asadi-2017,Rubin-2012,Fox-2016,Kappen-2012,Theodorou-2012,Theodorou-2013} are globally optimal in that they minimize expected path cost subject to a fixed total relative entropy of paths constraint (see Equation (\ref{Eq_optimization_problem_BoP01})), at least in our setting of a absorbing, goal, node $n_{\mathcal{S}}$ reachable from any other node of the graph.

Interestingly, from Equations (\ref{Eq_real_expected_cost01}) and (\ref{Eq_expected_entropy03}), the cost function (\ref{Eq_optimization_problem_BoP01}) can be rewritten at the local, edge, level as $\sum_{i,j \in \mathcal{V} \setminus n} \bar{n}_{ij} c_{ij} + T \sum_{i \in \mathcal{V} \setminus n} \bar{n}_{i} \sum_{j \in \mathcal{S}ucc(i)} p_{ij} \log ( p_{ij}/p^{\mathrm{ref}}_{ij} )$ where $\bar{n}_{ij}$ is the expected flow through edge $(i,j)$ and $\bar{n}_{i}$ the expected number of visits to $i$ (see \cite{Akamatsu-1997,Bavaud-2012,Guex-2015} and \cite{Saerens-2008}, section 6.2). In this expression, the entropy term defined on each node is weighted by the expected number of visits to the node. The policy can thus also be obtained by minimizing this ``local" cost function in function of the transition probabilities defined on unconstrained nodes.

\subsubsection{The standard value iteration algorithm}

Let us first recall the standard value iteration procedure, computing the expected cost until absorption by the goal state $n_{\mathcal{S}}$ \cite{Puterman-1994,Sutton-2017,Bertsekas-1998,Bertsekas-2000} when starting from a state $k \in \mathcal{S}$, denoted by $v_{k}$, based on the following recurrence formula verified at optimality
\begin{equation}
v_{k} =
  \begin{cases}
   { \displaystyle \min_{a \in \mathcal{A}(k)} } \Bigg \{ c_{ka} + {\displaystyle \sum_{l \in \mathcal{S}ucc(a)}} p^{\mathrm{ref}}_{al} v_{l} \Bigg \} & \text{if } k \in \mathcal{S} \setminus \{ n_{\mathcal{S}} \} \\
   \phantom{-} 0       & \text{if } k = n_{\mathcal{S}}
  \end{cases}
\label{Eq_value_iteration01}
\end{equation}
where $v_{k}$ is the value (expected cost) from state $k$ and $p^{\mathrm{ref}}_{al}$ is element $a,l$ (with $a \in \mathcal{A}$ and $l \in \mathcal{S}$) of the transition matrix of the reference random walk on the bipartite graph.
This expression is iterated until convergence, which is guaranteed under some mild conditions, for any set of nonnegative initial values (see, e.g., \cite{Powell-2011,Puterman-1994,Sutton-2017,Bertsekas-1998,Bertsekas-2000} for details).

\subsubsection{The soft value iteration algorithm}

Let us start from the standard softmin-based expression computing the free energy directed distance in a regular graph (Equation (\ref{Eq_potential_recurrence_formula01}); see also \cite{Francoisse-2013,Francoisse-2017,Fouss-2016}). We observe that it corresponds to the Bellman-Ford expression providing the shortest-path distance in which the min operator has been replaced by the softmin operator defined in Equation (\ref{Eq_softmin01}).

Substituting in the same way the min operator (\ref{Eq_softmin01}) for the softmin, with the $p_{ka}^{\mathrm{ref}}$ playing the role of the weighting factors $q_{i}$, in the value iteration update formula (\ref{Eq_value_iteration01}) provides a ``soft" equivalent of the Bellman-Ford optimality conditions on the set of state nodes $\mathcal{S}$,
\begin{equation}
\phi_{k}^{\mathcal{S}} =
  \begin{cases}
   -\frac{1}{\theta} \log \left[ {\displaystyle \sum_{a \in \mathcal{A}(k)}} p_{ka}^{\mathrm{ref}} \exp \Bigg[-\theta \Big( c_{ka} + {\displaystyle \sum_{l \in \mathcal{S}ucc(a)}} p^\mathrm{ref}_{al} \phi_{l}^{\mathcal{S}} \Big) \Bigg] \right] & \text{if } k \in \mathcal{S} \setminus \{ n_{\mathcal{S}} \} \\
   \phantom{-} 0       & \text{if } k = n_{\mathcal{S}}
  \end{cases}
\label{Eq_value_iteration_MDP01}
\end{equation}
In the case of our bipartite graph of Figure \ref{Fig_bipartiteGraph01}, this equation can exactly be obtained by applying the recurrence expression computing the free energy in the constrained RSP (Equation (\ref{Eq_value_iteration_constrained_RSP01})), after recalling that the cost of the transition between an action node and a state node is equal to zero. More precisely, we simply substitute $\phi_{j}^{*}$ in the first line of Equation (\ref{Eq_value_iteration_constrained_RSP01}) by the expression in the second line, $\phi_{j}^{*} = $ $\sum_{l \in \mathcal{S}ucc(j)} p_{jl}^{\mathrm{ref}} \big( 0 + \phi^{*}_{l} \big)$, which directly provides Equation (\ref{Eq_value_iteration_MDP01}). Recall that the $p^{\mathrm{ref}}_{ka}$, $k \in \mathcal{S}$ and $a \in \mathcal{A}(k)$, correspond to the reference, prior, policy commonly set to a uniform distribution on the possible actions in state $k$, $p^{\mathrm{ref}}_{ka} = 1/|\mathcal{A}(k)|$. Conversely, the $p^{\mathrm{ref}}_{ak}$ with $a \in \mathcal{A}$ and $k \in \mathcal{S}$ are provided by the environment.

Note that it can easily be shown by following the same reasoning as in the appendix of \cite{Francoisse-2013,Francoisse-2017} that this recurrence formula reduces to the standard optimality conditions for Markov decision processes (Equation (\ref{Eq_value_iteration01})) when $\theta \rightarrow \infty$. Conversely, when $\theta \rightarrow 0^{+}$, it reduces to the expression allowing to compute the expected cost until absorption by the goal state $n_{\mathcal{S}}$, also called the average first-passage cost \cite{Kemeny-1960,Norris-1997,Taylor-1998}, $\phi_{k} = \sum_{a \in \mathcal{A}(k)} p_{ka}^{\mathrm{ref}} ( c_{ka} + \sum_{l \in \mathcal{S}ucc(a)} p^\mathrm{ref}_{al} \phi_{l} )$.

The idea is to iterate (\ref{Eq_value_iteration_MDP01}) until convergence of the free energies to a fixed point where the optimality conditions (\ref{Eq_value_iteration_MDP01}) are verified (no change occurs any more). The procedure converges to a unique solution as it corresponds to a particular case of the iterative procedure for solving the constrained RSP (Equation (\ref{Eq_value_iteration_constrained_RSP01})); see Appendix \ref{Appendix2} for the proof. Then, the optimal policy for each state $k \in \mathcal{S}$, $k \ne n_{\mathcal{S}}$, is computed thanks to Equation (\ref{Eq_biased_transition_probabilities02}), which provides the probability of choosing action $a$ within state $k$.

This procedure, involving the iteration of Equation (\ref{Eq_value_iteration_MDP01}) and the computation of the optimal policy from Equation (\ref{Eq_biased_transition_probabilities02}), will be called the \textbf{soft value iteration} algorithm. As already stated, such soft variants of value iteration already appeared in control \cite{Todorov-2006} and exploration strategies for which an additional Kullback-Leibler cost term is incorporated in the immediate cost \cite{Rubin-2012,Fox-2016,Kappen-2012,Theodorou-2012,Theodorou-2013,Azar-2011,Azar-2012}. It was also recently proposed as an operator guiding exploration in reinforcement learning, and more specifically for the SARSA algorithm in \cite{Asadi-2016,Asadi-2017}. The present work therefore provides a new interpretation to this exploration strategy.
We apply this algorithm in the next Section \ref{Sec_experiments01} in order to solve simple Markov decision problems, for illustration.

\subsection{Markov decision processes with discounting}
\label{Sec_discouted_cost01}

Finaly, let us briefly discuss the concept of MDP with discounting. In this setting, we still consider the random walk on the graph $G_{\mathrm{b}}$ with reference transition probabilities $\mathbf{P}_{\mathrm{ref}}$. However, in contrast with our previous setting, here, no goal node is defined -- the Markov chain defining the random walk is regular. In addition, a \textbf{discounting factor} $\gamma \in \ ] 0,1 [$ is introduced to decrease the impact of future costs with time \cite{Sutton-2017}: immediate costs are more important than postponed ones.

In standard MDPs, the introduction of the discounting factor can be interpreted from two different points of view:
\begin{itemize}
\item each future cost, for instance appearing at time step $t$, is reduced by a factor $\gamma^{t}$.
\item at each time step, the random walker has a small chance $(1 - \gamma)$ of quitting the process (the contract is cancelled, the agent is killed, etc).
\end{itemize}
In the case of standard MDPs, these two interpretations lead to the same model; however, in the RSP framework, they take distinct forms. They are left for further work, but we quickly introduce the intuition behind them.

The first interpretation leads to a new soft value iteration expression that has to be iterated for paths with increasing length. This can be done by unfolding the network in time and then apply the RSP on this new directed acyclic graph, as described in \cite{Mantrach-2011}.
For the second interpretation, the problem can be tackled by introducing a cemetery node (a killing, absorbing state). The agent then has a $(1 - \gamma)$ probability of being teleported to this cemetery state with a zero cost after choosing any action. The soft value iteration expression (\ref{Eq_value_iteration_MDP01}) can be adapted to this new setting.
These two RSP with discounting models will be investigated in further work.

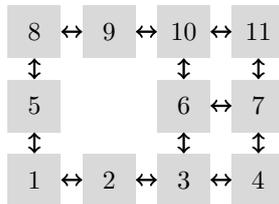
\begin{figure}
\begin{adjustbox}{width=0.3\textwidth,center}
			\begin{tikzpicture}[scale=1.5, transform shape]
				\tikzstyle{every node} = [rectangle, fill=gray!30, minimum height=20, minimum width=20]
				\node (1)  at (1, 0)  {1};
				\node (2)  at (2, 0)  {2};
				\node (3)  at (3, 0)  {3};
				\node (4)  at (4, 0)  {4};
				\node (5)  at (1, 1)  {5};
				\node[rectangle, fill=white, minimum width=20] (6)  at (2, 1)  { };
				\node (7)  at (3, 1)  {6};
				\node (8)  at (4, 1)  {7};
				\node (9)  at (1, 2)  {8};
				\node (10) at (2, 2)  {9};
				\node (11) at (3, 2) {10};
				\node (12) at (4, 2) {11};
				\foreach \from/\to in {	1/2, 1/5, 2/3, 3/4, 3/7, 4/8, 5/9, 7/8, 7/11, 8/12, 9/10, 10/11, 11/12}
				\draw [<->,line width=1] (\from) -- (\to);

			\end{tikzpicture}
\end{adjustbox}
			\caption{The maze problem. The goal of the agent is to reach node 11 from node 1. Notice that some transitions with no resulting displacement are possible (example in node 1: going west or south). The costs related to the actions are detailed in the text.}
				\label{Maze}
\end{figure}

\section{Some simulations: a simple illustration on the maze problem}
\label{Sec_experiments01}

This section illustrates the application of constrained randomized shortest paths to Markov decision problems. Several simulations were run on four different problems \cite{Lebichot-2018} but, in order to save space and because the conclusions are similar, we decided to report only one simple application: the probabilistic maze game inspired by \cite{Russell-2003}, as illustrated in Figure \ref{Maze}.

\begin{figure}
\center
\includegraphics[width=0.6\textwidth]{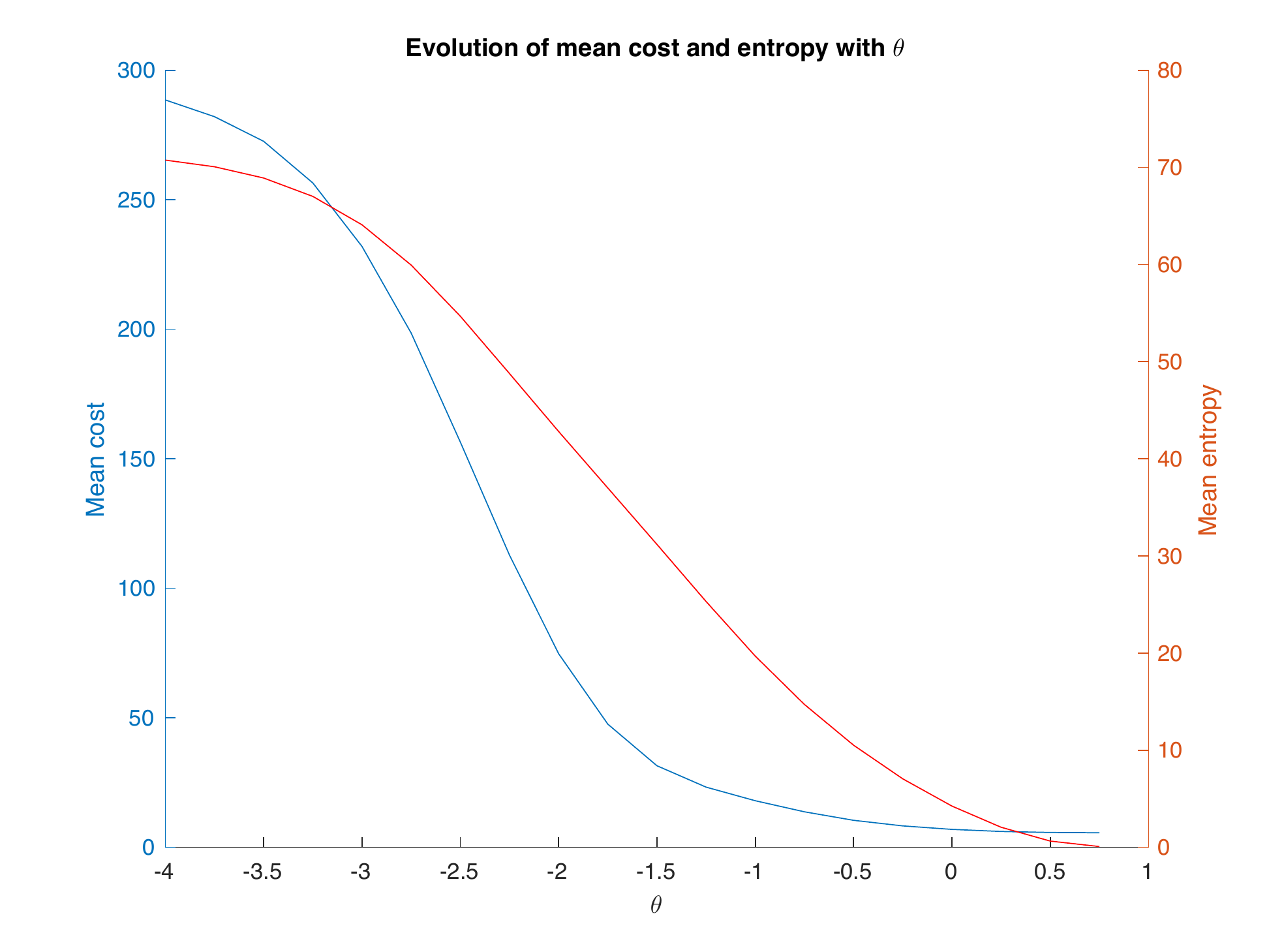}
\caption{Results, averaged over $10^{6}$ runs, obtained by simulating the policy provided by the constrained RSP when increasing $\theta$ (in $\log$ scale). The blue curve depicts the evolution of the average cost (mean number of turns to reach square 11 -- the smaller the best) in function of $\theta$. The red curve indicates the corresponding entropy of the state nodes (entropy of the randomized policies). Naturally, the largest entropy and average cost are achieved when $\theta$ is small and are minimum when $\theta$ is large.}
\label{Mazethetavsturn}      
\end{figure}

\begin{figure}
\centering
\begin{subfigure}[]{0.60\textwidth}
      \includegraphics[width = \textwidth, clip=true]{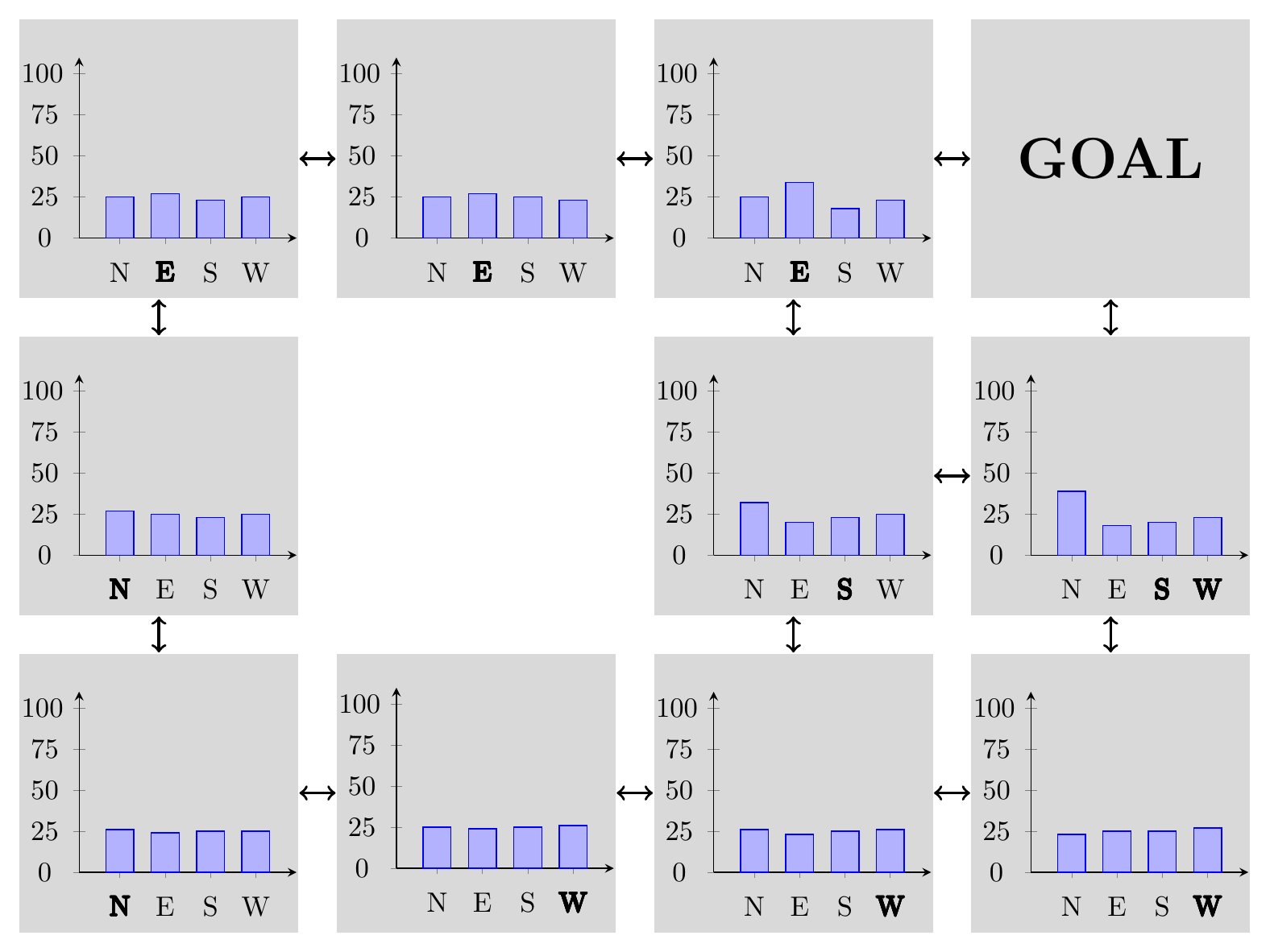}
      \caption{$\uparrow$ Inverse temperature $\theta=10^{-2.5}$.}
\end{subfigure}
\qquad
\begin{subfigure}[]{0.60\textwidth}
      \includegraphics[width = \textwidth, clip=true]{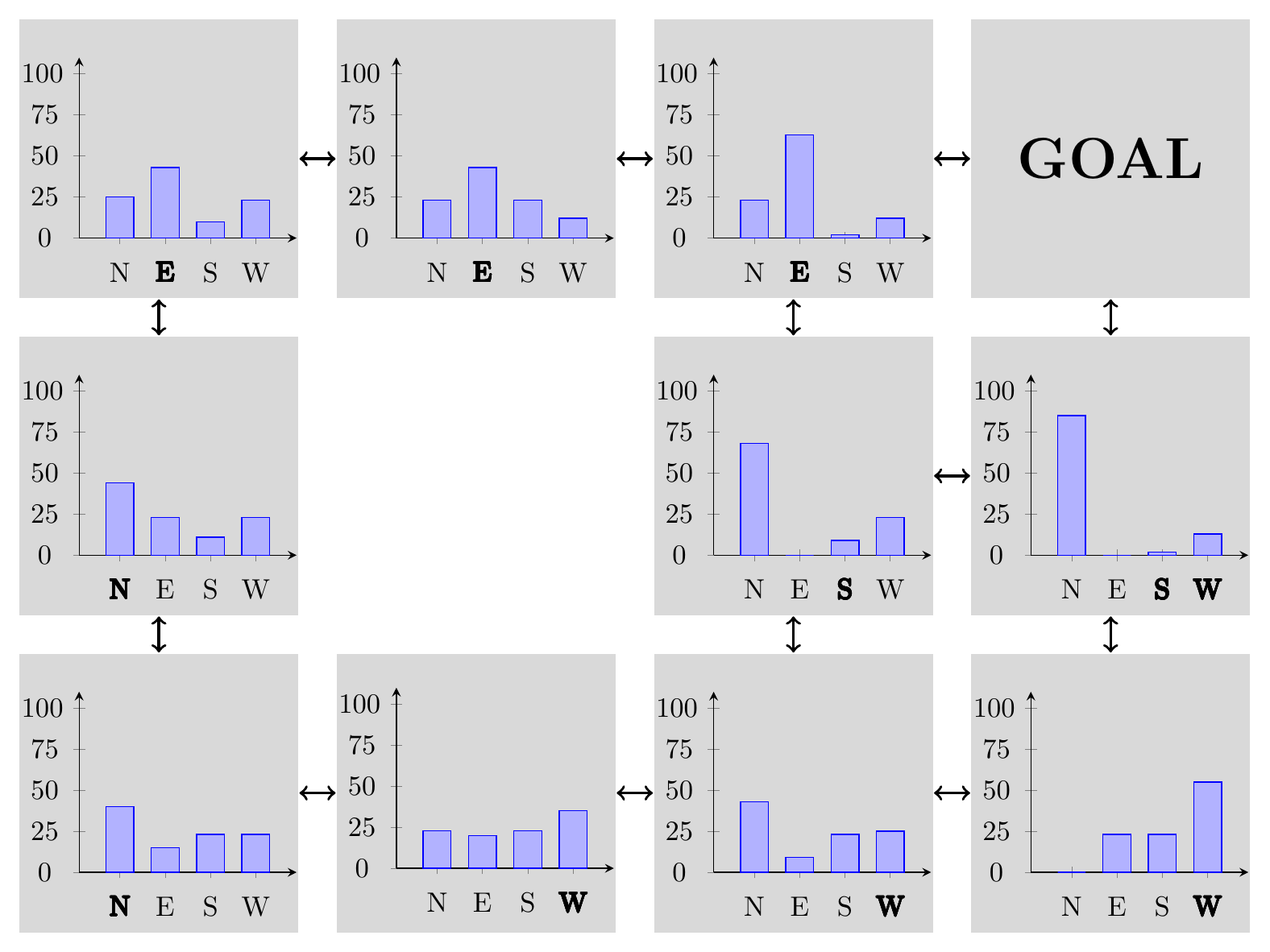}
      \caption{$\uparrow$ Inverse temperature $\theta=10^{-1}$.}
\end{subfigure}
\begin{subfigure}[]{0.60\textwidth}
      \includegraphics[width = \textwidth, clip=true]{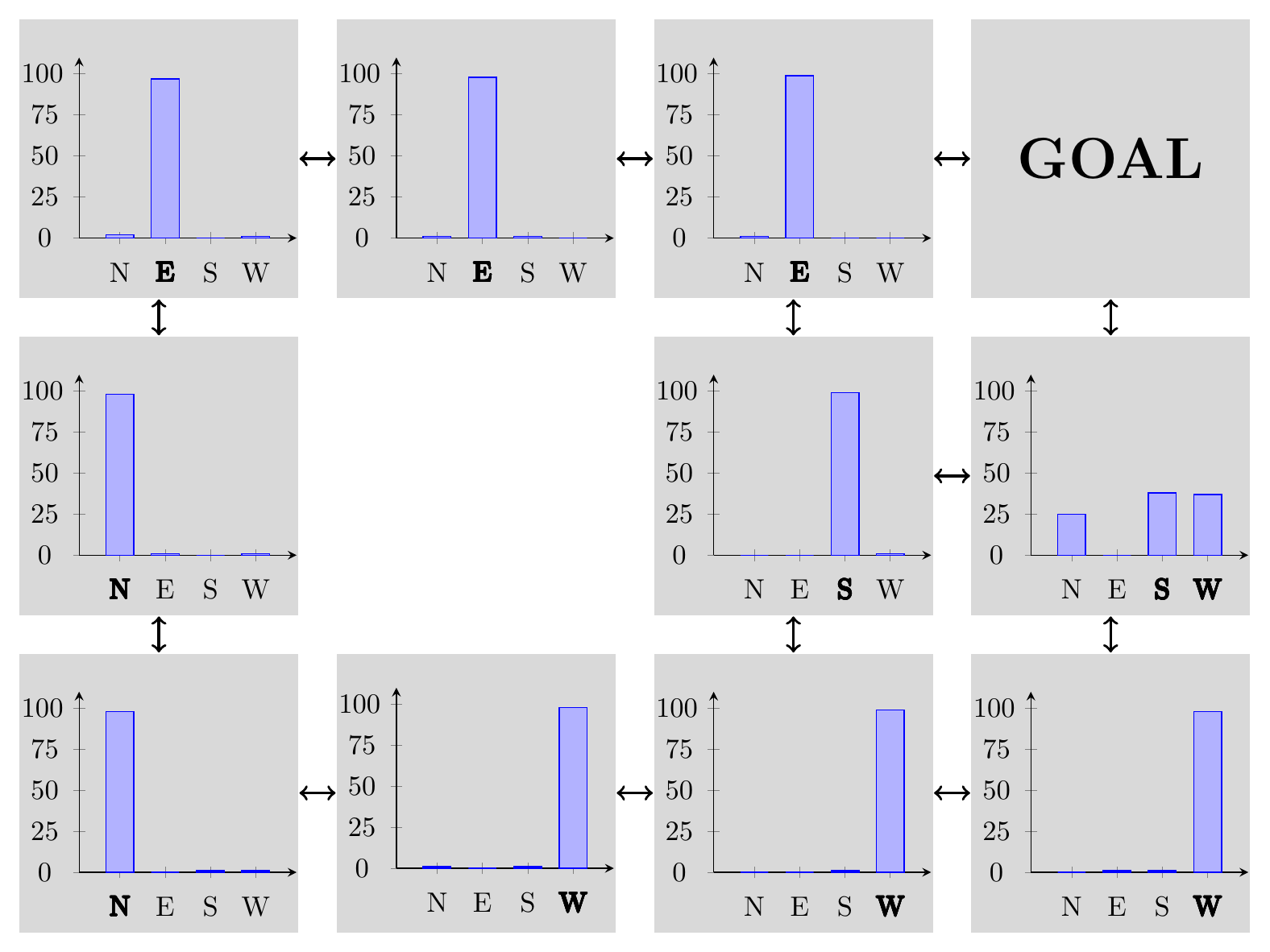} 
      \caption{$\uparrow$ Inverse temperature $\theta=10^{+0.5}$.}
\end{subfigure}
\caption{Optimal randomized policy obtained after convergence of the soft value iteration algorithm for three different, increasing, values of the inverse temperature parameter $\theta$. In each square, the agent has to choose between going north (N), east (E), south (S) and west (W). A larger $\theta$ corresponds to a more deterministic policy. Note that the optimal deterministic policy is indicated in bold in each case.}
\label{Fig_policies01}
\end{figure}

An agent, initially starting on square 1, is asked to reach square 11 in a minimum number of time steps (see Figure \ref{Maze}) and incurs some additional costs described below. To do so, the agent can choose between four actions in each square:
\begin{itemize}
\item Go north. However, a bug (for instance due to adverse wind conditions) can occur when the agent decides to go north so that it has only a $0.8$ probability to actually go north (no bug). When this bug occurs ($0.2$ probability), it then has a $0.5$ probability to go east (globally, a 10\% chance) and a $0.5$ probability to go west (also a 10\% chance globally).
\item Go east, west or south with probability one.
\end{itemize}
A unit cost is associated to each time step and, in addition, the cost for visiting square 7 is $+100$. This implies that if the above mentioned bug occurs on square 7 and the agent is redirected to east, the cost is increased again by $+100$ because the agent re-enters square 7.
Indeed, if the agent selects a direction which leads nowhere (a wall), for example selecting ``go east" or ``go west" in square 5, it stays in its current position, but incurs (again) the cost associated to the current square. 

Concerning the reference probabilities $p^{\mathrm{ref}}_{ij}$, these are defined by the environment on action nodes and are set to $1/4$ on state nodes (a purely random policy).

Note that the optimal policy from square 1 is $1 \rightarrow 5 \rightarrow 8 \rightarrow 9 \rightarrow 10 \rightarrow 11$. We ran a large number of simulations of the process until the agent reaches the goal node (each simulation from initial state 1 to goal state 11 is called a run), for a range of randomized policies (depending on the parameter $\theta$) obtained after running the soft value iteration algorithm (\ref{Eq_value_iteration_MDP01}).

Figure \ref{Mazethetavsturn} represents the evolution of the mean cost to reach goal node 11 and the entropy (computed only on state nodes, thus on the randomized policies) in function of $\theta$. The results are averaged over $10^{6}$ runs for each value of $\theta$, with the policy obtained after convergence of the soft value iteration algorithm for this $\theta$ (see Equation (\ref{Eq_value_iteration_MDP01})). We observe that the largest average cost and entropy are achieved when $\theta$ is small, and are smallest when $\theta$ is large. The resulting functions are both logistic-shaped between two bounds:
\begin{itemize}
\item When $\theta$ is small, entropy is maximum as each action has approximately a $1/4$ probability to be chosen and therefore the expected scores are the same as for a pure random walk.
\item When $\theta$ becomes large, entropy and expected cost (as well as the policy) are almost the same as for the standard value iteration algorithm providing the optimal deterministic policy.
\end{itemize}

Moreover, the optimal randomized policies obtained after convergence of the soft value iteration algorithm for three different, increasing, values of the inverse temperature parameter $\theta$ are illustrated in Figure \ref{Fig_policies01}.
This example clearly shows that using a randomized strategy allows to balance the strength of the player.

\section{Conclusion}
\label{Sec_conclusion01}

This work presented two procedures for solving constrained randomized shortest-paths problems, together with an application to randomized Markov decision processes where the problem is viewed as a bipartite graph. The main objective is to reach a goal node from an initial node in a graph while minimizing expected cost subject to a relative entropy equality constraint and transition probabilities constraints on some edges. The model provides a randomized policy encouraging exploration, balancing exploitation and exploration. The amount of exploration is monitored by the inverse temperature parameter.

The problem is expressed in terms of full paths connecting the initial node to the goal node and can easily be solved. The solution is a Gibbs-Boltzmann probability distribution on the set of paths with virtual extra costs associated to the constrained edges.

Two algorithms for computing the local policy at the edge level are developed. The first algorithm is based on Lagrange duality and requires solving iteratively the standard randomized shortest-paths problem until convergence.
The second algorithm is reminiscent of Bellman-Ford's algorithm for solving the shortest-path distance problem. It simply aims to replace the min operator by a softmin operator in Bellman-Ford's recurrence relation to update the expected cost on unconstrained nodes. For the constrained nodes, because the transition probabilities are fixed, we simply use the expression for computing the expected cost until absorption in a Markov chain. The convergence of the procedure is guaranteed for the two algorithms.

The usefulness of these algorithms is then illustrated on standard Markov decision problems. Indeed, a standard Markov decision process can be reinterpreted as a randomized shortest-paths problem on a bipartite graph. Standard Markov decision problems can thus easily be solved by the two introduced algorithms: they provide a randomized policy minimizing expected cost under entropy and transition probabilities constraints.

This shows that the exploration strategy using the softmin instead of the min in the value iteration algorithm is optimal in the predefined sense. Therefore it justifies the previous work \cite{Asadi-2016,Asadi-2017,Azar-2011,Azar-2012,Fox-2016,Kappen-2012,Rubin-2012,Theodorou-2012,Theodorou-2013,Todorov-2006} from a randomized shortest-paths point of view.

Future work will focus on extending the randomized shortest-paths model in order to deal with other types of constraints. In particular we will work on inequality constraints on transition probabilities, as well as flow equality and inequality constraints, on both node flows and edge flows. Another interesting extension of the randomized shortest-paths model is the multi-sources multi-destinations randomized optimal transport on a graph generalizing the deterministic optimal transport on a graph problem. We also plan to investigate constrained randomized shortest paths with a discounting factor as well as average-reward Markov decision processes which were recently studied in the light of entropy regularization \cite{Neu-2017}.

\section*{Acknowledgements}

This work was partially supported by the Immediate and the Brufence projects funded by InnovIris (Brussels region), as well as former projects funded by the Walloon region, Belgium. We thank these institutions for giving us the opportunity to conduct both fundamental and applied research.

We also thank Benjamin Blaise, a former Master student, who helped us to investigate the randomized Markov decision processes during his masters thesis at UCLouvain \cite{Blaise-2013}, as well as Prof. Fabrice Rossi for his useful remarks.

\begin{center}
\rule{2.5in}{0.01in} 
\par\end{center}

\appendix
\numberwithin{equation}{section}

\section*{Appendix: Additional material and proofs of the main results}
\label{Appendix}

\section[Appen000]{Computing quantities of interest}
\label{Ap_quantitiesOfInterest01}

In this appendix, several important quantities derived from the standard randomized shortest-paths (RSP) framework, and which will be needed in the paper, are detailled. The material is mainly taken from the previous work \cite{Yen-08K,Saerens-2008,Kivimaki-2012,Francoisse-2013,Francoisse-2017,Fouss-2016}.

\paragraph{The minimum free energy.}

Interestingly, if we replace the probability distribution $\mathbb{P}$ by the optimal distribution $\mathbb{P}^{*}$ provided by Equation (\ref{Eq_Boltzmann_probability_distribution01}) in the objective function (\ref{Eq_optimization_problem_BoP01}), we obtain for the minimum \textbf{free energy} between node $1$ and node $n$
\begin{align}
\phi_{1}^{*}(T) = \phi(\mathbb{P}^{*}) &= \dsum_{\wp \in \mathcal{P}} \mathrm{P}^{*}(\wp) \tilde{c}(\wp) + T \dsum_{\wp \in \mathcal{P}} \mathrm{P}^{*}(\wp) \log \left( \frac{\mathrm{P}^{*}(\wp)}{\tilde{\pi}(\wp)} \right) \nonumber \\
 &= \dsum_{\wp \in \mathcal{P}} \mathrm{P}^{*}(\wp) \tilde{c}(\wp) + T \dsum_{\wp \in \mathcal{P}} \mathrm{P}^{*}(\wp) \left( -\tfrac{1}{T} \tilde{c}(\wp) - \log \mathcal{Z} \right) \nonumber \\
 &= -T \log \mathcal{Z} = -\tfrac{1}{\theta} \log \mathcal{Z}
\label{Eq_optimal_free_energy01}
\end{align}

\paragraph{The expected number of passages through edges.}
For the \emph{expected number of passages through an edge} $(i,j)$ at temperature $T = 1/\theta$, that is, the flow in $(i,j)$, we obtain from last result (\ref{Eq_optimal_free_energy01}) and the definition of the partition function $\mathcal{Z}$ (Equation (\ref{Eq_Boltzmann_probability_distribution01})),
\begin{align}
\frac{\partial \phi^{*}_{1}}{\partial c_{ij}} &= \frac{\partial (-\tfrac{1}{\theta} \log \mathcal{Z})}{\partial c_{ij}} = - \frac{1}{\theta \mathcal{Z}} \frac{\partial \mathcal{Z}} {\partial c_{ij}}
 =  - \frac{1}{\theta \mathcal{Z}} \sum_{\wp\in\mathcal{P}} \tilde{\pi} (\wp)\exp[-\theta \tilde{c}(\wp)] (- \theta) \frac{\partial \tilde{c}(\wp)} {\partial c_{ij}} \nonumber \\
&= \sum_{\wp\in\mathcal{P}} \frac{ \tilde{\pi}(\wp) \exp[-\theta \tilde{c}(\wp)] } {\mathcal{Z}} \frac{\partial \tilde{c}(\wp)} {\partial c_{ij}}
= \sum_{\wp\in\mathcal{P}} \mathrm{P}^{*}(\wp) \, \eta\big((i,j) \in \wp\big)
\triangleq \bar{n}_{ij}(T)
\label{Eq_expected_visits_edges}
\end{align}
where we used $\partial \tilde{c}(\wp)/\partial c_{ij} = \eta\big((i,j) \in \wp\big)$, with $\eta\big((i,j) \in \wp\big)$ being the number of times edge $(i,j)$ appears on path $\wp$ at temperature $T$. Therefore, we have for the flow in $(i,j)$
\begin{equation}
\bar{n}_{ij}(T) = -T \frac{\partial \log\mathcal{Z}} {\partial c_{ij}}
\label{Eq_flows_from_partition_function01}
\end{equation}

\paragraph{Computation of the partition function.}
Now, it turns out that the partition function can easily be computed in closed form (see, e.g., \cite{Saerens-2008,Kivimaki-2014,Fouss-2016} for details). Let us first introduce the \textbf{fundamental matrix} of the RSP system,
\begin{equation}
\mathbf{Z} = \mathbf{I} + \mathbf{W} + \mathbf{W}^{2} + \cdots = (\mathbf{I} - \mathbf{W})^{-1}, \quad \text{with } \mathbf{W} = \mathbf{P}_{\mathrm{ref}} \circ \exp[-\theta \mathbf{C}]
\label{Eq_fundamentalMatrix01}
\end{equation}
where $\mathbf{C}$, $\mathbf{P}_{\mathrm{ref}}$ are respectively the cost and the reference transition probabilities matrices (see Equation (\ref{Eq_referenceRandomWalk01})) while $\circ$ is the elementwise (Hadamard) product. Elementwise, the entries of the $\mathbf{W}$ matrix are $w_{ij} = [\mathbf{W}]_{ij} = p^{\mathrm{ref}}_{ij} \exp[-\theta c_{ij}]$, except for the goal node where $w_{nj} = 0$ for all $j$ (killing, absorbing, node). Note that this matrix is sub-stochastic because the costs are non-negative and node $n$ is absorbing and killing (row $n$ contains only 0 values).

Then, the partition function is simply $\mathcal{Z} = [\mathbf{Z}]_{1n} = z_{1n}$ (see \cite{Yen-08K,Saerens-2008,Kivimaki-2012,Francoisse-2013,Francoisse-2017}). More generally \cite{Garcia-Diez-2011,Francoisse-2013}, it can be shown that the elements $z_{in}$ of the fundamental matrix correspond to
\begin{equation}
z_{in} = \dsum_{\wp \in \mathcal{P}_{in}} \tilde{\pi}(\wp) \exp[-\theta \tilde{c}(\wp)]
\label{Eq_forward_backward_variables01}
\end{equation}
with $z_{nn} = 1$, and where $\mathcal{P}_{in}$ is the set of hitting paths starting in node $i$ and ending in killing absorbing node $n$. The $z_{in}$ quantities are usually called the backward variables. They can be interpreted as probabilities of surviving during a killed random walk with transition matrix $\mathbf{W}$, that is, reaching hitting node $n$ without being killed during the walk (see, e.g., \cite{Francoisse-2013,Francoisse-2017} for details).

\paragraph{Computation of the expected number of passages and visits.}
Moreover, the flow in $(i,j)$ can be obtained from (\ref{Eq_fundamentalMatrix01}) and the expression $\partial \mathbf{M}^{-1} / \partial x = - \mathbf{M}^{-1} (\partial \mathbf{M} / \partial x) \mathbf{M}^{-1}$ (see, e.g., \cite{Harville-97}),
\begin{equation}
\bar{n}_{ij}(T) = - \tfrac{1}{\theta} \frac{\partial \log\mathcal{Z}} {\partial c_{ij}}
= \frac{ z_{1i} p_{ij}^{\mathrm{ref}} \exp[-\theta c_{ij}]  z_{jn} } {z_{1n}}
= \frac{ z_{1i} w_{ij}  z_{jn} } {z_{1n}}
\label{Eq_computation_edge_flows01}
\end{equation}
and because only the first row and the last column of $\mathbf{Z}$ are needed, two systems of linear equations can be solved instead of matrix inversion in Equation (\ref{Eq_fundamentalMatrix01}).

From the last equation and $z_{in} = \sum_{j=1}^{n} w_{ij} z_{jn} + \delta_{in}$ (the elementwise form of $(\mathbf{I} - \mathbf{W}) \mathbf{Z} = \mathbf{I}$), the \emph{expected number of visits to a node} $j$ can be computed from
\begin{equation}
\bar{n}_{i}(T) \triangleq \sum_{j=1}^{n} \bar{n}_{ij}(T) + \delta_{in} = \frac{ z_{1i} z_{in} } {z_{1n}} \quad \text{for } i \ne n
\label{Eq_computation_node_flows01}
\end{equation}
where we assume $i \ne n$ for the last equality because we already know that $\bar{n}_{n}(T) = 1$ at the goal node, which is absorbing and killing.


\paragraph{The optimal randomized policy.}
Furthermore, from (\ref{Eq_computation_edge_flows01})-(\ref{Eq_computation_node_flows01}), the optimal transition probabilities of following an edge $(i,j)$ with $i \ne n$ are
\begin{equation}
p^{*}_{ij}(T) = \frac{\bar{n}_{ij}(T)}{\bar{n}_{i}(T)}
= p_{ij}^{\mathrm{ref}} \exp[-\theta c_{ij}] \frac{z_{jn}}{z_{in}}
= \frac{w_{ij} z_{jn}}{z_{in}}
= \frac{w_{ij} z_{jn}}{\sum_{j'= \mathcal{S}ucc(i)} w_{ij'} z_{j'n}}
\label{Eq_biased_transition_probabilities01}
\end{equation}
because $p_{ij}^{\mathrm{ref}} \exp[-\theta c_{ij}] = w_{ij}$ and $z_{in} = \sum_{j = \mathcal{S}ucc(i)} w_{ij} z_{jn}$  for all $i \ne n$ (the elementwise form of $(\mathbf{I} - \mathbf{W}) \mathbf{Z} = \mathbf{I}$, coming from Equation (\ref{Eq_fundamentalMatrix01})). 
This expression defines a \textbf{biased random walk} on $G$ -- the random walker is ``attracted" by the goal node $n$. These transition probabilities define a first-order Markov chain and do not depend on the source node. They correspond to the optimal, randomized, ``routing" strategy, or policy, minimizing free energy from the current node. This policy will therefore be called the \textbf{randomized policy} in the sequel (a mixed policy or strategy in game theory \cite{Osborne-2004}).

\paragraph{The expected cost until destination.}
In addition, the expected cost until reaching goal node $n$ from node 1 is \cite{Saerens-2008,Kivimaki-2014,Fouss-2016}
\begin{align}
  \langle \tilde{c} \rangle = \dsum_{\wp \in \mathcal{P}} \mathrm{P}^{*} (\wp) \tilde{c}(\wp)
  = \dsum_{\wp \in \mathcal{P}} \frac{\tilde{\pi}(\wp) \exp[-\theta \tilde{c}(\wp)]}{\mathcal{Z}} \tilde{c}(\wp)
  \label{Eq_expected_cost01}
\end{align}
 
After defining the matrix containing the expected number of passages through the edges by $\mathbf{N}$ with $[\mathbf{N}]_{ij} = \bar{n}_{ij}(T)$, it can be shown by proceeding in the same way as for Equation (\ref{Eq_expected_visits_edges}) (see \cite{Saerens-2008} for details) that the expected cost spent in the network is
\begin{equation}
\langle \tilde{c} \rangle = - \frac{\partial \log\mathcal{Z}} {\partial \theta} = \mathbf{e}^{\text{T}} (\mathbf{N} \circ \mathbf{C}) \mathbf{e}
\label{Eq_real_expected_cost01}
\end{equation}
where $\mathbf{e}$ is a column vector of 1s and $\circ$ is the elementwise (Hadamard) matrix product.
This quantity is just the cumulative sum of the expected number of passages through each edge times the cost of following the edge, $\sum_{i=1}^{n-1} \sum_{j \in \mathcal{S}ucc(i)}  \bar{n}_{ij}(T) c_{ij}$ \cite{Garcia-Diez-2011}.

%

\paragraph{The entropy of the paths.}

In Equation (\ref{Eq_optimization_problem_BoP01}), the relative entropy of the set of paths, for the optimal probability distribution, was defined as
\begin{equation}
J(\mathbb{P}^{*}|\tilde{\pi}) = \dsum_{\wp \in \mathcal{P}} \mathrm{P}^{*}(\wp) \log \left( \frac{\mathrm{P}^{*}(\wp)}{\tilde{\pi}(\wp)} \right)
\label{Eq_expected_entropy01}
\end{equation}
and, from Equations (\ref{Eq_optimization_problem_BoP01}), (\ref{Eq_optimal_free_energy01}) and (\ref{Eq_real_expected_cost01}), can be computed thanks to
\begin{equation}
J(\mathbb{P}^{*}|\tilde{\pi})
= -( \log \mathcal{Z} + \tfrac{1}{T} \langle \tilde{c} \rangle )
\label{Eq_expected_entropy02}
\end{equation}
where the partition function $\mathcal{Z} = [\mathbf{Z}]_{1n} = z_{1n}$.

In addition, it can be shown that the total entropy of the set of paths is \cite{Akamatsu-1997,Saerens-2008}
\begin{equation}
J(\mathbb{P}^{*}) = - \dsum_{\wp \in \mathcal{P}} \mathrm{P}^{*}(\wp) \log \mathrm{P}^{*}(\wp)
= - \sum_{i=1}^{n-1} \bar{n}_{i} \sum_{j \in \mathcal{S}ucc (i)} p^{*}_{ij}(T) \log( p^{*}_{ij}(T) )
\label{Eq_expected_entropy03}
\end{equation}
which sums the local entropies over the transient (non-absorbing) nodes weighted by the expected number of visits to each node.

\paragraph{The free energy distance.}
It was already shown in Equation (\ref{Eq_optimal_free_energy01}) that the minimal free energy (\ref{Eq_optimal_free_energy01}) at temperature $T$ is provided by
\begin{equation}
\phi_{1}^{*}(T) = \phi(\mathbb{P}^{*}) = -T \log \mathcal{Z} = -\tfrac{1}{\theta} \log z_{1n}
\label{Eq_free_energy_definition01}
\end{equation}
In \cite{Francoisse-2013,Francoisse-2017}, it was proved that the free energy from any starting node $i$ to absorbing, goal, node $n$, $\phi_{i}^{*}(T) = -\tfrac{1}{\theta} \log z_{in}$, can be computed thanks to the following recurrence formula to be iterated until convergence
\begin{equation}
\phi_{i}^{*}(T) =
  \begin{cases}
   -\frac{1}{\theta} \log \left[ {\displaystyle \sum_{j \in \mathcal{S}ucc(i)}} p_{ij}^{\mathrm{ref}} \exp[-\theta (c_{ij} + \phi_{j}^{*}(T))] \right] & \text{if } i \ne n \\
   \phantom{-} 0       & \text{if } i=n
  \end{cases}
\label{Eq_potential_recurrence_formula01}
\end{equation}
This equation is an extension of Bellman-Ford's formula for computing the shortest-path distance in a graph (see, e.g., \cite{Bertsekas-2000,Christofides_1975,Cormen-2009,Gondran-1984,Jungnickel-2005,Rardin-1998,Sedgewick-2011}). Moreover, the recurrence expression (\ref{Eq_potential_recurrence_formula01}) is also a generalization of the distributed consensus algorithm developed in \cite{Tahbaz-2006}, considering binary costs only.

It was also shown \cite{Francoisse-2013,Francoisse-2017} that this minimal free energy interpolates between the least cost ($T = \theta^{-1} \rightarrow \infty$; $\phi_{i}^{*}(\infty) = \min_{j \in \mathcal{S}ucc(i)} \{ c_{ij} + \phi_{j}^{*}(\infty) \}$ and $\phi_{n}^{*}(\infty) = 0$) and the expected cost before absorption ($T = \theta^{-1} \rightarrow 0^{+}$; $\phi_{i}^{*}(0) = \sum_{j \in \mathcal{S}ucc(i)} p_{ij}^{\mathrm{ref}} ( c_{ij} + \phi_{j}^{*}(0) )$ and $\phi_{n}^{*}(0) = 0$) \cite{Kivimaki-2012,Francoisse-2013,Francoisse-2017}.
In addition, this quantity defines a \textbf{directed distance} between any node and absorbing node $n$ \cite{Kivimaki-2012,Francoisse-2013,Francoisse-2017}. This directed free energy distance has a nice interpretation: it corresponds (up to a scaling factor) to minus the logarithm of the probability of reaching node $n$ without being killed during a killed random walk defined by the sub-stochastic transition probabilities $w_{ij} = p^{\mathrm{ref}}_{ij} \exp[-\theta c_{ij}]$ \cite{Francoisse-2013,Francoisse-2017}. In other words, it is minus the logarithm of the probability of \emph{surviving} during the walk. Still another interesting result is that, when computing the continuous time -- continuous state equivalent to the RSP model by densifying the graph, the free energy becomes a \emph{potential} attracting the agents to the goal state \cite{Garcia-Diez-2011b}.

\paragraph{The softmin operator.}
In fact, as discussed in \cite{Francoisse-2013,Francoisse-2017}, this last expression (\ref{Eq_potential_recurrence_formula01}) is obtained by simply substituting the $\min$ operator by a weighted version of the \textbf{softmin operator} (\cite{Cook-2011}; also called the \textbf{log-sum-exp} function \cite{BoydBook-2004,Murphy-2012,Tahbaz-2006}) in the Bellman-Ford recurrence formula,
\begin{equation}
\operatorname{softmin}_{\mathbf{q},\theta}(\mathbf{x}) = -\tfrac{1}{\theta} \log\bigg( \sum_{j=1}^{n} q_{j} \exp[-\theta x_{j}] \bigg), \text{ with all } q_{j} \ge 0 \text{ and } {\textstyle \sum_{j=1}^{n} q_{j}=1}
\label{Eq_softmin01}
\end{equation}
which is a smooth approximation of the min operator and interpolates between weighted average and minimum operators, depending on the parameter $\theta$ \cite{Cook-2011,Tahbaz-2006}.
This expression also appeared in control \cite{Todorov-2006} and exploration strategies for which an additional Kullback-Leibler cost term is incorporated in the immediate cost \cite{Rubin-2012,Fox-2016,Kappen-2012,Theodorou-2012,Theodorou-2013,Azar-2011,Azar-2012}. Moreover, this function\footnote{They actually study the softmax counterpart.} was recently proposed as an operator guiding exploration in reinforcement learning, and more specifically for the SARSA algorithm \cite{Asadi-2016,Asadi-2017} -- see these references for a discussion of its properties.

\paragraph{The randomized policy in terms of free energy.}
Note that the optimal randomized policy derived in Equation (\ref{Eq_biased_transition_probabilities01}) can be rewritten in function of the free energy as
\begin{equation}
p^{*}_{ij}(T) = \frac{p_{ij}^{\mathrm{ref}} \exp[-\theta c_{ij}] z_{jn}}{\sum_{j'=1}^{n} p_{ij'}^{\mathrm{ref}} \exp[-\theta c_{ij'}] z_{j'n}}
= \frac{p_{ij}^{\mathrm{ref}} \exp[-\theta (c_{ij} + \phi^{*}_{j}(T))]}{\sum_{j'=1}^{n} p_{ij'}^{\mathrm{ref}} \exp[-\theta (c_{ij'} + \phi^{*}_{j'}(T))]}
\label{Eq_biased_transition_probabilities02}
\end{equation}
because $z_{in} = \exp[-\theta \phi^{*}_{i}(T)]$ and $z_{in} = \sum_{j=1}^{n} w_{ij} z_{jn}$ $ = $ $\sum_{j=1}^{n} p_{ij}^{\mathrm{ref}} \exp[-\theta c_{ij}] z_{jn}$ for all $i \ne n$. This corresponds to a multinomial logistic function.


\section[Appen00]{Solving the system of logistic equations \label{Appendix0}}

In this appendix, we are mainly interested in deriving the solution of a simple system of multinomial logistic equations. Assume we have to solve the following equations
\begin{equation}
\frac{\gamma_{i} \exp[-\theta x_{i}]}{\sum_{j=1}^{n} \gamma_{j} \exp[-\theta x_{j}]} = q_{i} \quad \text{with each } q_{i}, \gamma_{i} \ge 0
\label{Eq_logistic_equation01}
\end{equation}
with respect o the $x_{i}$, together with the following equality constraints
\begin{equation}
  \begin{cases}
      \sum_{i=1}^{n} q_{i} = 1\\
      \sum_{i=1}^{n} q_{i} x_{i} = 0
  \end{cases}
  \label{Eq_constraints_x_problem01}
\end{equation}

The multinomial logistic function in (\ref{Eq_logistic_equation01}) is often encountered in applied statistics, for instance it forms the main functional form of the multivariate logistic model \cite{Hosmer-2000}. In this appendix, we derive the solution $\mathbf{x}^{*}$ of this equation satisfying the given constraints and then use it in order to solve Equation (\ref{Eq_to_solve_augmented_costs01}). The second equality constraint in (\ref{Eq_constraints_x_problem01}) is introduced because any shift of a solution vector, $\mathbf{x}^{*} - \mathbf{c}$, is also a solution. Adding this second constraint solves the problem of degeneracy.

Taking the ratio between the two equations (\ref{Eq_logistic_equation01}) involving $q_{i}$ and $q_{j}$ and taking $- \tfrac{1}{\theta} \log$ of both sides gives $x_{i} - x_{j} = -\tfrac{1}{\theta} [\log(q_{i}/\gamma_{i}) - \log(q_{j}/\gamma_{j})]$. This provides $n-1$ independent equations and a common practice is to set one value to $0$, for instance $x_{n} = 0$ \cite{Hosmer-2000}. Here, we will instead force the second equality constraint (\ref{Eq_constraints_x_problem01}). Multiplying both sides by $q_{j}$ and summing over $j$ provides $x_{i} - \sum_{j=1}^{n} q_{j} x_{j} = -\tfrac{1}{\theta} [\log(q_{i}/\gamma_{i}) - \sum_{j=1}^{n} q_{j} \log(q_{j}/\gamma_{j})]$ (recall that the $q_{i}$ sum to $1$ and $\sum_{i=1}^{n} q_{i} x_{i} = 0$) gives
\begin{equation}
x_{i} = -\tfrac{1}{\theta} \Big( \log(q_{i}/\gamma_{i}) - \sum_{j=1}^{n} q_{j} \log(q_{j}/\gamma_{j}) \Big)
\label{Eq_expression_for_x01}
\end{equation}


%

We now apply this result in order to solve Equation (\ref{Eq_to_solve_augmented_costs01}) with $x_{j} = \myDelta_{ij}$ (we condition the computation on an arbitrary node $i$). By comparing (\ref{Eq_to_solve_augmented_costs01}) with (\ref{Eq_logistic_equation01}) as well as recalling that $p^{\mathrm{ref}}_{ij} = q_{ij}$ and $\phi^{*}_{i} = -\tfrac{1}{\theta} \log z_{in}$ (Equation (\ref{Eq_minimum_free_energy01})), we observe that $\gamma_{j} = p^{\mathrm{ref}}_{ij} \exp[- \theta c_{ij}] z_{jn}$ and therefore $-\tfrac{1}{\theta} \log(q_{j}/\gamma_{j}) = \tfrac{1}{\theta} \log( \exp[-\theta c_{ij}] z_{jn} ) = -(c_{ij} + \phi^{*}_{j})$. Injecting this result in (\ref{Eq_expression_for_x01}) finally provides for constrained nodes
\begin{equation}
\myDelta_{ij} = -(c_{ij} + \phi^{*}_{j}) + \sum_{k \in \mathcal{S}ucc(i))} p^{\mathrm{ref}}_{ik} (c_{ik} + \phi^{*}_{k})
\label{Eq_my_delta_expression_appendix01}
\end{equation}
which is the require result.

\section[Appen01]{Derivation of the iterative algorithm \label{Appendix1}}

In order to compute the optimal policy $p^{*}_{ij}$, we observe from Equation (\ref{Eq_biased_transition_probabilities02}) that we need to find the free energy $\phi^{*}_{j} = - \log z_{jn}$, and thus the backward variable $z_{jn}$ starting from a node $j$,
\begin{equation}
p^{*}_{ij} \propto  p_{ij}^{\mathrm{ref}} \exp[-\theta (c_{ij} + \phi^{*}_{j})] \nonumber
\label{Eq_optimal_policy_proportional_to01}
\end{equation}
where $\propto$ means ``proportional to".
The quantity $p^{*}_{ij}$ then needs to be normalized so that $\sum_{j \in \mathcal{S}ucc(i)} p^{*}_{ij} = 1$.
We will therefore have to compute the backward variable $z_{jn}$ for the two sets of nodes of interest, the constrained nodes $\mathcal{C}$ and the unconstrained nodes $\mathcal{U}$.

From the definition of the backward variable (Equation (\ref{Eq_forward_backward_variables01}), but now including the augmented costs on constrained nodes), we obtain by decomposing the paths $i \leadsto n$ into the first step $i \rightarrow j$, and then the remaining steps $j \leadsto n$ (see \cite{Garcia-Diez-2011} for a related derivation),
\begin{align}
z_{in} &= \dsum_{\wp_{in} \in \mathcal{P}_{in}} \tilde{\pi}(\wp_{in}) \exp[-\theta \tilde{c}'(\wp_{in})] \nonumber \\
&= \dsum_{j \in \mathcal{S}ucc(i)} \dsum_{\wp_{jn} \in \mathcal{P}_{jn}} p^{\mathrm{ref}}_{ij} \tilde{\pi}(\wp_{jn}) \exp[-\theta (c'_{ij} + \tilde{c}(\wp_{jn}))] \nonumber \\
&= \dsum_{j \in \mathcal{S}ucc(i)} p^{\mathrm{ref}}_{ij} \exp[-\theta c'_{ij}]  \underbracket[0.5pt][5pt]{ \dsum_{\wp_{jn} \in \mathcal{P}_{jn}} \tilde{\pi}(\wp_{jn}) \exp[-\theta \tilde{c}'(\wp_{jn})] }_{z_{jn}} \nonumber \\
&= \dsum_{j \in \mathcal{S}ucc(i)} p^{\mathrm{ref}}_{ij} \exp[-\theta c'_{ij}]  z_{jn}
\label{Eq_FE_augmented_costs_bipartite01}
\end{align}
where $\wp_{in}$ is a path starting in a node $i$ and ending in the killing, absorbing, node $n$. We will now express this recurrence formula in terms of the free energy, which will lead to an interesting extension of the Bellman-Ford formula.

Taking $- \frac{1}{\theta} \log$ of this last expression and recalling that $\phi^{*}_{i} = - \frac{1}{\theta} \log z_{in}$ yields, for any node $i \ne n$,
\begin{equation}
\phi^{*}_{i} = - \tfrac{1}{\theta} \log \dsum_{j \in \mathcal{S}ucc(i)} p^{\mathrm{ref}}_{ij} \exp[-\theta (c'_{ij} + \phi^{*}_{j})]
\label{Eq_Bellman_Ford_general01}
\end{equation}

The remainder of the development depends on the type of node $i$; we therefore continue with the unconstrained nodes, followed by the constrained ones.

\subsection{Computation of the free energy on unconstrained nodes}

For unconstrained nodes, $c'_{ij} = c_{ij}$ and we simply have 
\begin{equation}
\phi^{*}_{i} = - \tfrac{1}{\theta} \log \dsum_{j \in \mathcal{S}ucc(i)} p^{\mathrm{ref}}_{ij} \exp[-\theta (c_{ij} + \phi^{*}_{j})] \quad \text{for each } i \in \mathcal{U}
\label{Eq_backward_variable_states01}
\end{equation}
because there is no augmented cost associated to the transitions from an unconstrained node -- they are not part of the set of constrained transitions (see Subsection \ref{Subsec_basic_MDP_model01}). This corresponds to the standard recurrence formula for computing the free energy in the RSP framework (see Equation (\ref{Eq_potential_recurrence_formula01}) or \cite{Francoisse-2013,Francoisse-2017}).
Let us now compute this quantity on constrained nodes.

\subsection{Computation of the free energy on constrained nodes}
\label{App_subsec_freeEnergy_constrained01}

In the case of constrained nodes, we have to use the augmented costs $c'_{ij}$ in order to ensure that the relative flow in the edge $(i,j)$ is equal to the predefined transition probability $p_{ij}^{\mathrm{ref}}$ provided by the environment. Remember that the value of these augmented costs can be expressed in function of the free energy, $c'_{ij} = \sum_{l \in \mathcal{S}ucc(i)} p_{il}^{\mathrm{ref}} ( c_{il} + \phi^{*}_{l} ) - \phi^{*}_{j}$ (Equation (\ref{Eq_formula_augmented_costs_free_energy01})).
Injecting this result in Equation (\ref{Eq_Bellman_Ford_general01}) provides
\begin{align}
\phi^{*}_{i} &= - \tfrac{1}{\theta} \log \dsum_{j \in \mathcal{S}ucc(i)} p^{\mathrm{ref}}_{ij} \exp[-\theta (c'_{ij} + \phi^{*}_{j})] \nonumber \\
&= - \tfrac{1}{\theta} \log \dsum_{j \in \mathcal{S}ucc(i)} p^{\mathrm{ref}}_{ij} \exp \Bigg[-\theta \Big( \sum_{l \in \mathcal{S}ucc(i)} p_{il}^{\mathrm{ref}} ( c_{il} + \phi^{*}_{l} ) \Big) \Bigg] \nonumber \\
&= - \tfrac{1}{\theta} \log  \Bigg[  \exp \Bigg[-\theta \Big( \sum_{l \in \mathcal{S}ucc(i)} p_{il}^{\mathrm{ref}} ( c_{il} + \phi^{*}_{l} ) \Big) \Bigg]  \Bigg( \dsum_{j \in \mathcal{S}ucc(i)} p^{\mathrm{ref}}_{ij} \Bigg) \Bigg] \nonumber \\
&= \sum_{l \in \mathcal{S}ucc(i)} p_{il}^{\mathrm{ref}} ( c_{il} + \phi^{*}_{l} ), \quad \text{for each } i \in \mathcal{C}
\label{Eq_backward_variable_actions01}
\end{align}
Moreover, for the goal node $n$, $z_{nn}=1$ so that $\phi^{*}_{n}=0$. This last result as well as Equations (\ref{Eq_backward_variable_states01})-(\ref{Eq_backward_variable_actions01}) therefore justify the recurrence formula (\ref{Eq_value_iteration_constrained_RSP01}).

\section[Appen02]{Convergence of the iterative algorithm \label{Appendix12}}
 \label{Appendix2}

In this appendix, the convergence of the iteration algorithm based on Equation (\ref{Eq_value_iteration_constrained_RSP01}) is shown based on the fixed point theorem.

First, let us observe that the solution to the recurrence relation (\ref{Eq_value_iteration_constrained_RSP01}) (two first lines of the equation) is invariant up to a translation of the origin of the free energy. Indeed, it can easily be shown that if $\boldsymbol{\phi}^{*}$ is a solution of (\ref{Eq_value_iteration_constrained_RSP01}), a shift of the free energy by a quantity $\alpha$, that is ${{\phi}_{i}^{*}}' = \phi_{i}^{*} + \alpha$ for each $i$, is also a solution to (\ref{Eq_value_iteration_constrained_RSP01}). To overcome this underdetermination, the free energy is set to zero on the absorbing, goal, node $n$, $\phi_{n}^{*} = 0$.

We will now study the following fixed point iteration after permuting the index of the nodes so that the unconstrained nodes appear before the constrained nodes,
\begin{equation}
\phi^{*}_{i} \leftarrow
  \begin{cases}
   -\frac{1}{\theta} \log \left[ {\displaystyle \sum_{j=1}^{n}} p_{ij}^{\mathrm{ref}} \exp \Big[-\theta \Big( c_{ij} + \phi^{*}_{j} \Big) \Big] \right] & \text{if } 1 \le i \le |\mathcal{U}| \\
   {\displaystyle \phantom{-} \sum_{j=1}^{n}} p_{ij}^{\mathrm{ref}} \big( c_{ij} + \phi^{*}_{j} \big) & \text{if } |\mathcal{U}| + 1 \le i \le |\mathcal{U}| + |\mathcal{C}| \\
   \phantom{-} 0       & \text{if } i=n
  \end{cases}
\label{Eq_value_iteration_constrained_restricted_RSP01}
\end{equation}

Then, it is well-known that this kind of fixed-point iteration converges to a unique solution in a convex domain (here, the positive quadrant) if the Jacobian matrix, $\mathbf{J}$, of the transformation has a matrix norm (for instance its spectral radius) strictly smaller than 1 everywhere in this domain \cite{Dahlquist-1974,Phillips-1996}. In that case, the fixed-point transformation is what is called a contraction mapping. We will thus compute the spectral radius of the Jacobian matrix and verify that it is smaller than one for all non-negative values of $\boldsymbol{\phi}^{*}$.

The element $i,j$ of this Jacobian matrix can easily be computed from Equation (\ref{Eq_value_iteration_constrained_restricted_RSP01}). For unconstrained nodes,
\begin{equation}
[\mathbf{J}]_{ij} = \frac{\partial \phi_{i}^{*}} {\partial \phi_{j}^{*}}
= \frac{ p_{ij}^{\mathrm{ref}} \exp \Big[-\theta \Big( c_{ij} + \phi^{*}_{j} \Big) \Big] }
{ {\displaystyle \sum_{k=1}^{n}} p_{ik}^{\mathrm{ref}} \exp \Big[-\theta \Big( c_{ik} + \phi^{*}_{k} \Big) \Big] } \quad \text{for } 1 \le i \le |\mathcal{U}|
\label{Eq_jacobian_matrix01}
\end{equation}
For constrained nodes,
\begin{equation}
\frac{\partial \phi_{i}^{*}} {\partial \phi_{j}^{*}}
= p_{ij}^{\mathrm{ref}} \quad \text{for } |\mathcal{U}| + 1 \le i \le |\mathcal{U}| + |\mathcal{C}|
\label{Eq_jacobian_matrix02}
\end{equation}
and of course $\partial \phi_{n}^{*} / \partial \phi_{j}^{*} = 0$ for all $j$.

Then, we can verify that this Jacobian matrix $\textbf{J}$ is sub-stochastic. Indeed, row sums are equal to 1 for rows 1 to $(n-1)$, and the last row sum (for node $n$) is strictly less that 1 (it is equal to 0). Consequently, because, in addition, all the elements of the matrix are non-negative, $\mathbf{J}$ is sub-stochastic \cite{Meyer-2000}. Thus, $\textbf{J}$ defines a transition probability matrix of a killing, absorbing, Markov chain with a killing absorbing node $n$ \cite{Fouss-2016}.

Now, from the definition of the Jacobian matrix (\ref{Eq_jacobian_matrix01})-(\ref{Eq_jacobian_matrix02}), the graph induced by $\mathbf{J}$ connects the $n$ nodes in exactly the same way as the original graph $G$: node $i$ and node $j$ are connected if and only if they are connected in the original graph (the connectivity pattern is induced by $p^{\mathrm{ref}}_{ij}$).

Moreover, as it is assumed that the original graph $G$ is strongly connected, the absorbing, killing, node $n$ can be reached from any initial node of the graph and this property is kept for $\mathbf{J}$. This means that, exactly as in the case of a standard absorbing Markov chain, the total probability mass in the transient states of the network (nodes $1$ to $n-1$) will gradually decrease and $\lim_{t \rightarrow \infty} \mathbf{J}^{t} \rightarrow 0$ \cite{Grinstead-1997}. This implies that the spectral radius of the Jacobian matix $\mathbf{J}$ is strictly less than 1 \cite{Meyer-2000}. Therefore, as the spectral radius is a matrix norm, the iteration (\ref{Eq_value_iteration_constrained_restricted_RSP01}) converges to a unique solution independently of the (positive) initial conditions \cite{Dahlquist-1974,Phillips-1996}.


\begin{center}
\rule{2.5in}{0.01in} 
\par\end{center}

{\small 
\bibliographystyle{abbrv}
\bibliography{randomizedMDP}
 }{\small \par}
 
\begin{center}
\rule{2.5in}{0.01in} 
\par\end{center}

\end{document}